\def\year{2021}\relax
\documentclass[letterpaper]{article} 
\usepackage{aaai21}  
\usepackage{times}  
\usepackage{helvet} 
\usepackage{courier}  
\usepackage[hyphens]{url}  
\usepackage{graphicx} 
\usepackage{natbib}  
\usepackage{caption} 
\usepackage{amssymb}
\usepackage{amsmath}
\usepackage{booktabs}
\usepackage{multirow}
\usepackage{makecell}
\usepackage{subfigure}
\usepackage{pdfpages}
\DeclareMathOperator*{\argmax}{argmax}

\newtheorem{theorem}{Theorem}
\newtheorem{definition}{Definition}
\newcommand{\ie}{\emph{i.e.}}
\newcommand{\resp}{\emph{resp.}}
\newcommand{\etal}{\emph{et\ al.}}

\title{Disentangled Information Bottleneck}
\author{
    Ziqi Pan, Li Niu,\thanks{Corresponding author} Jianfu Zhang, Liqing Zhang\footnotemark[1]\\
}
\affiliations{
    MoE Key Lab of Artificial Intelligence, Department of Computer Science and Engineering\\
    Shanghai Jiao Tong University, Shanghai, China\\
    \{panziqi\_ai, ustcnewly, c.sis\}@sjtu.edu.cn, zhang-lq@cs.sjtu.edu.cn\\
}

\begin{document}

\maketitle

\begin{abstract}
    The information bottleneck (IB) method is a technique for extracting information that is relevant for predicting the target random variable from the source random variable, which is typically implemented by optimizing the IB Lagrangian that balances the compression and prediction terms. However, the IB Lagrangian is hard to optimize, and multiple trials for tuning values of Lagrangian multiplier are required. Moreover, we show that the prediction performance strictly decreases as the compression gets stronger during optimizing the IB Lagrangian. In this paper, we implement the IB method from the perspective of supervised disentangling. Specifically, we introduce \emph{Disentangled Information Bottleneck} (DisenIB) that is consistent on compressing source maximally without target prediction performance loss (maximum compression). Theoretical and experimental results demonstrate that our method is consistent on maximum compression, and performs well in terms of generalization, robustness to adversarial attack, out-of-distribution detection, and supervised disentangling.
\end{abstract}

\section{Introduction}

Compression is a ubiquitous task in machine learning~\cite{cover2012elements,mackay2003information}. For example, over-parameterized deep networks are compressed with pruning for the sake of computational efficiency~\cite{han2015deep,dai2018compressing}, machines may transform complex data into compressed representations that generalizes well~\cite{alemi2016deep,shwartz2017opening}. It is important to determine which aspects of data should be preserved and which should be discarded. The \emph{Information Bottleneck}~\cite{tishby2000information,dimitrov2001neural,samengo2002information} (IB) method provides a principled approach to this problem, which compresses the source random variable to keep the information relevant for predicting the target random variable while discarding all irrelevant information. The IB method has been applied to various domains such as classification~\cite{hecht2009speaker}, clustering~\cite{slonim2005information}, coding theory~\cite{hassanpour2018equivalence,zeitler2008design}, and quantization~\cite{strouse2017deterministic,cheng2019risk}. Recent research also demonstrate that the IB method can produce well-generalized representations~\cite{shamir2010learning,vera2018role,amjad2019learning} and may be promising on explaining the learning behaviors of neural networks~\cite{tishby2015deep,shwartz2017opening,saxe2019information}.

Given random variables $X,Y$ with their joint distribution $p_\mathrm{data}\left(X,Y\right)$, the IB method aims to compress $X$ to a ``bottleneck'' random variable $T$ keeping the information relevant for predicting $Y$. Namely, seeking a probabilistic mapping $q\left(T|X\right)$ such that the Mutual Information (MI) $I\left(X;T\right)$ is constrained while the information $I\left(T;Y\right)$ is maximized, which can be formally stated in terms of the constrained optimization problem
\begin{equation}\label{eq_ib_nonlinear_constrain}
    \mathop{\argmax}\limits_{T\in\Delta} I\left(T;Y\right),\quad\mathrm{s.t.}\ I\left(X;T\right)\leqslant r,
\end{equation}
where a level of compression (\ie, $I\left(X;T\right)\leqslant r$) is provided as the constraint, and $\Delta$ is the set of random variables $T$ that obey the Markov chain $Y\leftrightarrow X\leftrightarrow T$~\cite{witsenhausen1975conditional,ahlswede1975source,gilad2003information}. Optimal solutions to Eq.~\eqref{eq_ib_nonlinear_constrain} for $r\in\left[0,+\infty\right)$ form the \emph{IB curve}~\cite{tishby2000information,rodriguez2020convex}, which specifies the trade-off between the compression (\ie, $I\left(X;T\right)$) and prediction (\ie, $I\left(T;Y\right)$) terms. In practice, to avoid the non-linear constraint, Eq.~\eqref{eq_ib_nonlinear_constrain} can be optimized by minimizing the so-called \emph{IB Lagrangian}~\cite{tishby2000information,gilad2003information,shamir2010learning,rodriguez2020convex}
\begin{equation}\label{eq_ib_lagrangian}
    \mathcal{L}_\mathrm{IB}\left[q\left(T|X\right);\beta\right]=-I\left(T;Y\right)+\beta I\left(X;T\right),
\end{equation}
where $\beta$ is the Lagrange multiplier which controls the trade-off and typically set in $\left[0,1\right]$~\cite{rodriguez2020convex}.

Minimizing the IB Lagrangian encounters the following two problems: 1) It is hard to obtain the desired compression level $r$ (Eq.~\eqref{eq_ib_nonlinear_constrain}). In practice, a compression level is expected to obtain via minimizing the IB Lagrangian with certain $\beta$. However, recent works~\cite{kolchinsky2018caveats,rodriguez2020convex} pointed out that $\beta$ and the compression level are not causally related, and therefore multiple optimizations for different $\beta$ value are required to achieve a specific compression level. 2) $I\left(T;Y\right)$ given by the optimal solution to the IB Lagrangian is strictly decreasing as $\beta$ increases, \ie, the prediction performance is unavoidably reduced by the compression. We provide theoretical justification for this statement (see Theorem~\ref{theorem_trade_off}).

It is expected to extract the minimal sufficient~\cite{friedman2001elements} part about $Y$ from $X$ into $T$, \ie, compressing $X$ maximally without reducing $I\left(T;Y\right)$, which is referred to as \emph{maximum compression} in the remainder of this paper. However, such a case cannot be achieved through minimizing the IB Lagrangian, since compression always decreases $I\left(T;Y\right)$. Moreover, it is expected to eliminate the need for multiple optimizations and explore a consistent method for maximum compression with a single optimization. We start by realizing that \emph{supervised disentangling}~\cite{ridgeway2016survey} is closely related to the idea behind the IB method. Supervised disentangling tackles the problem of identifying complementary data aspects and separating them from each other with supervision. Similarly, in the IB method, one must separate $Y$-relevant and $Y$-irrelevant data aspects. This inspires us to implement the IB method from the perspective of supervised disentangling, leading to our proposed \emph{Disentangled Information Bottleneck} (DisenIB). To the best of our knowledge, we are the first to draw the connection between the IB method and supervised disentangling. Our contribution are threefold:
\begin{itemize}
  \item {
    We study the trade-off in the IB Lagrangian, showing that balancing compression and prediction terms can only decrease prediction performance, therefore the maximum compression can not be achieved.
  }
  \item {
    We propose a variant of IB, the \emph{Disentangled Information Bottleneck} (DisenIB), which is proven to be consistent on maximum compression. Specifically, DisenIB eliminates the need for multiple optimizations and consistently performs maximum compression with a single optimization.
  }
  \item {
    Through experimental results, we justify our theoretical statements and show that DisenIB performs well in terms of generalization~\cite{shamir2010learning}, robustness to adversarial attack~\cite{alemi2016deep} and out-of-distribution data detection~\cite{alemi2018uncertainty}, and supervised disentangling.
  }
\end{itemize}

The remainder of this paper is organized as follows. First, we analyze the trade-off in optimizing the IB Lagrangian in Section~\ref{sec_method_trade_off}, showing that the prediction performance strictly decreases as the compression gets stronger. To overcome the trade-off problem, we firstly propose a formal definition on compressing source data maximally without prediction performance loss (maximum compression) in Section~\ref{sec_method_consistency}, followed by introducing our proposed DisenIB that is consistent on maximum compression in Section~\ref{sec_method_disenib}. All our experimental analyses are provided in Section~\ref{sec_exp}.

\section{Methodology}\label{sec_method}

In this section, we first study the trade-off involved in the IB Lagrangian, showing that balancing compression and prediction terms can only decrease prediction performance and thus cannot achieve maximum compression. Then, we introduce our proposed DisenIB that is consistent on maximum compression.

\subsection{The IB Lagrangian Trade-off}\label{sec_method_trade_off}

We first show that optimizing the IB Lagrangian leads to inevitable trade-off. Specifically, the optimal solutions to the compression and prediction objectives obtained by optimizing the IB Lagrangian are consistently inferior to that obtained by optimizing each objective independently. This can be formally stated by the following Theorem~\ref{theorem_trade_off} (see supplementary for proof):
\begin{theorem}\label{theorem_trade_off}
    Consider the derivable IB Lagrangian,
    \begin{equation}\label{eq_trade_off_func}
        \mathcal{L}_\mathrm{IB}\left[q\left(T|X\right);\beta\right]=-I\left(T;Y\right)+\beta I\left(X;T\right),
    \end{equation}
    to be minimized over $q\left(T|X\right)$ with $\beta\geqslant0$. Let $q^*_\beta$ optimize $\mathcal{L}_\mathrm{IB}\left[q\left(T|X\right);\beta\right]$. Assume that $I_{q^*_\beta}\left(X;T\right)\not=0$,
    \begin{equation}\label{eq_trade_off_conc}
        \frac{\partial I_{q^*_\beta}\left(T;Y\right)}{\partial\beta}< 0\mathrm{\ and\ }
        \frac{\partial I_{q^*_\beta}\left(X;T\right)}{\partial\beta}< 0.
    \end{equation}
\end{theorem}
We can learn that for every nontrivial solution $q^*_\beta$ such that $I_{q^*_\beta}\left(X;T\right)\not=0$, $I\left(T;Y\right)$ strictly decreases as $\beta$ increases, and compression (\ie, $\beta I\left(X;T\right)$) can only decrease prediction performance (\ie, $I\left(T;Y\right)$), which is not expected.

\subsection{Consistency Property}\label{sec_method_consistency}

Optimizing the IB Lagrangian can not achieve maximum compression due to the aforementioned trade-off. It is expected to explore a method that is capable of performing maximum compression. Moreover, we also expect to eliminate the need for multiple optimizations. Namely, we expect to explore a method that consistently performs maximum compression with a single optimization, which is referred to as the \emph{consistency} property on maximum compression.

We first specify $r$ (Eq.~\eqref{eq_ib_nonlinear_constrain}) that provides the maximum compression case. We analyze the case where $Y$ is a deterministic function of $X$~\cite{rodriguez2020convex}, which covers a wide range of application scenarios such as classification and regression. In such a case, $I\left(X;Y\right)=H\left(Y\right)$. However, our method is also applicable in general cases where $I\left(X;Y\right)<H\left(Y\right)$. According to basic properties of MI~\cite{cover2012elements} (\ie, $I\left(T;Y\right)\leqslant H\left(Y\right)$ when $Y$ is discrete-valued), $I\left(T;Y\right)=H\left(Y\right)$ leads to the case without prediction loss. By leveraging the \emph{data processing inequality} (DPI)~\cite{cover2012elements} (\ie, $I\left(T;Y\right)\leqslant I\left(X;T\right)$) and basic properties of MI~\cite{cover2012elements} (\ie, $I\left(X;T\right)\leqslant H\left(X\right)$ when $X$ is discrete-valued), in the case without prediction loss, we have that
\begin{equation}\label{eq_maximum_ixt}
    H\left(Y\right)=I\left(T;Y\right)\leqslant I\left(X;T\right)\leqslant H\left(X\right).
\end{equation}
Hence $r=H\left(Y\right)$ provides the maximum compression, in which case
\begin{equation}\label{eq_maximum_cprs}
    I\left(X;T\right)=I\left(T;Y\right)=H\left(Y\right).
\end{equation}
In practice, we aim to design a cost function $\mathcal{L}$, such that the maximum compression case (Eq.~\eqref{eq_maximum_cprs}) is expected to be obtained via minimizing $\mathcal{L}$. Specifically, we expect that minimized $\mathcal{L}$ consistently satisfies Eq.~\eqref{eq_maximum_cprs}. Hence the formal definition of \emph{consistency} on maximum compression is given as
\begin{definition}[Consistency]\label{def_consistency}
    The lower-bounded cost functional $\mathcal{L}$ is consistent on maximum compression, if
    \begin{equation}
        \begin{split}
            & \forall\epsilon>0, \exists \delta>0,\quad\mathcal{L}-\mathcal{L}^*<\delta \Longrightarrow \\
            & \quad \left|I\left(X;T\right)-H\left(Y\right)\right|+\left|I\left(T;Y\right)-H\left(Y\right)\right|<\epsilon,
        \end{split}
    \end{equation}
    where $\mathcal{L}^*$ is the global minimum of $\mathcal{L}$.
\end{definition}
Satisfying Eq.~\eqref{eq_maximum_cprs} involves precise information amount control, \ie, exactly constraining both $I\left(X;T\right)$ and $I\left(T;Y\right)$ at $H\left(Y\right)$. Several works~\cite{alemi2016deep,chen2016infogan,kolchinsky2019nonlinear} involve estimating MI. However, they can only maximize or minimize MI, but still struggle to constrain the MI at an exact value. For the examples~\cite{alemi2016deep,kolchinsky2019nonlinear} using variational bounds, the estimation only becomes accurate as the bound becomes tight. There also exist MI estimators like MINE~\cite{belghazi2018mutual} that can provide accurate estimations, but certain sample complexity~\cite{belghazi2018mutual} is required. Therefore, they exhibit high variances~\cite{jiaming2019understanding}, and it is non-trivial to achieve Eq.~\eqref{eq_maximum_cprs}.

\subsection{Disentangled IB}\label{sec_method_disenib}

We introduce our proposed DisenIB that is consistent on maximum compression. After realizing the relation between IB and supervised disentangling, we implement IB from the perspective of supervised disentangling by introducing another random variable $S$ as the complementary aspect to $T$ and further encoding information relevant (\resp, irrelevant) to $Y$ into $T$ (\resp, $S$). Formally, the objective functional to be minimized is stated as
\begin{equation}\label{eq_disenib_obj}
    \begin{split}
        & \mathcal{L}_\mathrm{DisenIB}\left[q\left(S|X\right), q\left(T|X\right)\right] \\
        & \quad =-I\left(T;Y\right)-I\left(X;S,Y\right)+I\left(S;T\right).
    \end{split}
\end{equation}
Specifically, we encourage $\left(S,Y\right)$ to represent the overall information of $X$ by maximizing $I\left(X;S,Y\right)$, so that $S$ at least covers the information of $Y$-irrelevant data aspect. We encourage that $Y$ can be accurately decoded from $T$ by maximizing $I\left(T;Y\right)$, so that $T$ at least covers the information of $Y$-relevant data aspect. Hence, the amount of information stored in $S$ and $T$ are both lower bounded. In such a case, forcing $S$ to be disentangled from $T$ by minimizing $I\left(S;T\right)$ eliminates the overlapping information between them and thus tightens both bounds, leaving the exact information relevant (\resp, irrelevant) to $Y$ in $T$ (\resp, $S$).

Moreover, maximum compression can be consistently achieved via optimizing $\mathcal{L}_\mathrm{DisenIB}$, as stated in the following Theorem~\ref{prop_disenib_consistent} (see supplementary for proof):
\begin{theorem}\label{prop_disenib_consistent}
    $\mathcal{L}_\mathrm{DisenIB}$ is consistent on maximum compression.
\end{theorem}

We now introduce how to implement DisenIB in principle, leaving practical implementation in supplementary due to space limitation. Same as prior works~\cite{alemi2016deep,chalk2016relevant,achille2018information,kolchinsky2019nonlinear}, we derive variational approximations to $I\left(T;Y\right)$ and $I\left(X;S,Y\right)$ terms. By introducing variational probabilistic mappings $p\left(y|t\right)$ and $r\left(x|s,y\right)$, the tractable variational lower bounds can be formulated as
\begin{align}
      & \begin{aligned}
        I\left(T;Y\right) & = \mathbb{E}_{q\left(y,t\right)}\log q\left(y|t\right) - \mathbb{E}_{q\left(y\right)}\log q\left(y\right) \\
        & \geqslant \mathbb{E}_{q\left(y,t\right)}\log p\left(y|t\right) + H\left(Y\right),
      \end{aligned}  \label{eq_disenib_vlb_ity} \\
      & \begin{aligned}
        I\left(X;S,Y\right) & = \mathbb{E}_{q\left(x,s,y\right)}\log q\left(x|s,y\right) - \mathbb{E}_{q\left(x\right)}\log q\left(x\right) \\
        & \geqslant \mathbb{E}_{q\left(x,s,y\right)}\log r\left(x|s,y\right) + H\left(X\right),
      \end{aligned} \label{eq_disenib_vlb_ixsy}
\end{align}
where the inequalities follow from
\begin{align}
    & \begin{aligned}
        & \mathbb{E}_{q\left(y|t\right)}\log q\left(y|t\right) - \mathbb{E}_{q\left(y|t\right)}\log p\left(y|t\right) \\
        & \quad =D_\mathrm{KL}\left[q\left(Y|t\right)\parallel p\left(Y|t\right)\right]\geqslant 0,
    \end{aligned} \\
    & \begin{aligned}
        & \mathbb{E}_{q\left(x|s,y\right)}\log q\left(x|s,y\right) - \mathbb{E}_{q\left(x|s,y\right)}\log r\left(x|s,y\right) \\
        & \quad =D_\mathrm{KL}\left[q\left(X|s,y\right)\parallel r\left(X|s,y\right)\right]\geqslant 0.
    \end{aligned}
\end{align}
The lower bounds become tight as $D_\mathrm{KL}\left[q\left(Y|t\right)\parallel p\left(Y|t\right)\right]$ and $D_\mathrm{KL}\left[q\left(X|s,y\right)\parallel r\left(X|s,y\right)\right]$ approximate $0$. By rewriting (leveraging Markov chains $Y\leftrightarrow X\leftrightarrow T$ and $Y\leftrightarrow X\leftrightarrow S$)
\begin{align}
    & q\left(y,t\right) = \mathbb{E}_{p_\mathrm{data}\left(x\right)}p_\mathrm{data}\left(y|x\right)q\left(t|x\right), \\
    & q\left(x,s,y\right) = p_\mathrm{data}\left(x\right)p_\mathrm{data}\left(y|x\right)q\left(s|x\right),
\end{align}
we see that lower bounds in Eq~\eqref{eq_disenib_vlb_ity}-\eqref{eq_disenib_vlb_ixsy} only require samples from joint data distribution $p_\mathrm{data}\left(x,y\right)$, probabilistic mappings $q\left(t|x\right),q\left(s|x\right)$ and variational approximations $p\left(y|t\right),r\left(x|s,y\right)$, therefore is tractable.

Minimizing the $I\left(S;T\right)=D_\mathrm{KL}\left[q\left(S,T\right)\parallel q\left(S\right)q\left(T\right)\right]$ term is intractable since both $q\left(s,t\right)$ and $q\left(s\right)q\left(t\right)$ involve mixtures with a large number of components. However, we observe that we can sample from joint distribution $q\left(s,t\right)$ efficiently by first sampling $x$ from dataset uniformly at random and then sampling from $q\left(s,t|x\right)=q\left(s|x\right)q\left(t|x\right)$ due to the Markov chain $S\leftrightarrow X\leftrightarrow T$~\cite{kim2018disentangling}. We can also sample from product of marginal distributions $q\left(s\right)q\left(t\right)$ by shuffling the samples from the joint distribution $q\left(s,t\right)$ along the batch axis~\cite{belghazi2018mutual}. Then, we use the \emph{density-ratio-trick}~\cite{nguyen2008estimating,sugiyama2012density,kim2018disentangling} by involving a discriminator $d$ which estimates the probability that its input is a sample from $q\left(s,t\right)$ rather than from $q\left(s\right)q\left(t\right)$. Adversarial training is involved to train the discriminator,
\begin{equation}\label{eq_min_ist}
    \min\limits_{q}\max\limits_{d}
    \mathbb{E}_{q\left(s\right)q\left(t\right)}\log d\left(s,t\right)
    +\mathbb{E}_{q\left(s,t\right)}\log\left(1 - d\left(s,t\right)\right).
\end{equation}
As shown by~\cite{goodfellow2014generative}, $q\left(s,t\right)=q\left(s\right)q\left(t\right)$ when the Nash equilibrium is achieved, thus minimizing the $I\left(S;T\right)$ term.

\section{Relation to Prior Work}\label{sec_related}

In this section, we relate our proposed method to prior works on existing variants of IB Lagrangian and supervised disentangling methods.

\subsection{Variants of IB Lagrangian}

Optimizing the IB Lagrangian (Eq.~\eqref{eq_ib_lagrangian}) involves integrals which are typically intractable. For this reason, only limited cases~\cite{tishby2000information,chechik2005information} have been mainly developed until recently.

Recently, to optimize the IB Lagrangian on continuous and possibly non-Gaussian variables using neural networks, several works~\cite{alemi2016deep,chalk2016relevant,achille2018information,kolchinsky2019nonlinear} are proposed to derive variational approximations to the IB Lagrangian, which permits parameterizing the IB model using neural networks with gradient descent training. Specifically, they employed the same variational bound for the prediction term $I\left(T;Y\right)$ as our Eq.~\eqref{eq_disenib_vlb_ity}. However, they differ from ours in how to perform compression. While we compress via disentangling to avoid balancing compression and prediction terms, they all use variational upper bounds on the compression term $I\left(X;T\right)$. For VIB~\cite{alemi2016deep} and similar works~\cite{chalk2016relevant,achille2018information}, $I\left(X;T\right)$ is bounded as
\begin{equation}
    \begin{split}
        I\left(X;T\right) & = \mathbb{E}_{q\left(x,t\right)}\log q\left(t|x\right) - \mathbb{E}_{q\left(t\right)} \log q\left(t\right) \\
        & \leqslant \mathbb{E}_{q\left(x,t\right)}\log q\left(t|x\right) - \mathbb{E}_{q\left(t\right)} \log v\left(t\right),
    \end{split}
\end{equation}
where $v$ is some prior distribution. Differing from these three methods in upper bounding $I\left(X;T\right)$, the \emph{Nonlinear Information Bottleneck}~\cite{kolchinsky2019nonlinear} (NIB) uses a non-parametric upper bound based on \emph{Kernel Density Entropy Estimates}~\cite{kolchinsky2017estimating},
\begin{equation}\label{eq_nib_compression}
    I\left(X;T\right)\leqslant
    -\frac{1}{N}\sum_{i=1}^{N}\log\frac{1}{N}\sum_{j=1}^{N}
    e^{-D_\mathrm{KL}\left[q\left(t|x_i\right)\parallel q\left(t|x_j\right)\right]},
\end{equation}
where $N$ is the total number of samples in the given dataset.

Recent research~\cite{kolchinsky2018caveats} showed that optimizing the IB Lagrangian for different values of $\beta$ cannot explore the IB curve in the case where $Y$ is a deterministic function of $X$~\cite{rodriguez2019information}. To tackle this problem, they propose the \emph{squared-IB Lagrangian} by squaring $I\left(X;T\right)$:
\begin{equation}
    \mathcal{L}_\mathrm{squared-IB}=-I\left(T;Y\right)+\beta\left(I\left(X;T\right)\right)^2.
\end{equation}
G{\'a}lvez \etal~\cite{rodriguez2020convex} then took a further step to extend the squared-IB Lagrangian, showing that applying any monotonically increasing and strictly convex functions on $I\left(X;T\right)$ is able to explore the IB curve.

All these methods balance the compression and prediction terms. As we show in Theorem~\ref{theorem_trade_off}, IB methods involving such trade-off cannot achieve maximum compression. Differing from them, we alter compression to disentangling, which is shown to be able to avoid trade-off and consistently perform maximum compression.

\subsection{Supervised Disentangling}

\emph{Supervised disentangling}~\cite{ridgeway2016survey,fletcher1988robust} tackles the problem of identifying complementary aspects of data and separating them from each other with (partial) aspects label, which is a fundamental idea and various methods~\cite{mathieu2016disentangling,hadad2018two,jaiswal2018unsupervised,moyer2018invariant,zheng2019disentangling,song2019learning,gabbay2019demystifying,jaiswal2020invariant} are proposed. Though the IB method is closely related to supervised disentangling methods in the sense of separating target-relevant and target-irrelevant aspects of data, exactly controlling the amount of information stored in the respective data aspects is beyond the ability of these disentangling methods. Specifically, none of the optimality is guaranteed in the sense of information control for disentangling methods, which is referred to as \emph{information leak} in~\cite{gabbay2019demystifying}. For example~\cite{hadad2018two}, the redundant information irrelevant to prediction is compressed via limiting the expressive capacity of the neural network model, making it tricky to exactly control the amount of information. In~\cite{jaiswal2018unsupervised}, different terms are designed to compete with others, therefore tuning hyper-parameters that balance different terms is required, and exactly controlling information amount can not be achieved either. However, as we can see from the consistency property of our proposed DisenIB, our method can exactly control the amount of information stored in respective data aspects.

\section{Experiments}\label{sec_exp}

We compare existing variants of IB Lagrangian with our proposed DisenIB method. Following previous works~\cite{alemi2016deep,kolchinsky2019nonlinear,kolchinsky2018caveats,rodriguez2020convex}, we optimize existing methods for different values of $\beta\in\left[0, 1\right]$, producing a series of models that explore the trade-off between compression and prediction, while our method does not involve such trade-off. Following previous works~\cite{alemi2016deep,alemi2018uncertainty,kolchinsky2019nonlinear}, we evaluate our method in terms of generalization~\cite{shamir2010learning,vera2018role}, robustness to adversarial attack~\cite{alemi2016deep}, and out-of-distribution data detection~\cite{alemi2018uncertainty} on benchmark datasets: MNIST~\cite{lecun1998gradient}, FashionMNIST~\cite{xiao2017fashion}, and CIFAR10~\cite{krizhevsky2009learning}. We also provide results on more challenging natural image datasets: object-centric Tiny-ImageNet~\cite{deng2009imagenet} and scene-centric SUN-RGBD~\cite{song2015sun}. We also study the disentangling behavior of our method on MNIST~\cite{lecun1998gradient}, Sprites~\cite{reed2015deep} and dSprites~\cite{matthey2017dsprites}. Due to space limitation, the implementation details can be found in supplementary.

\subsection{Behavior on \emph{IB Plane}}\label{sec_exp_ib_trade_off}

\begin{figure*}[t]
  \centering
  \includegraphics[width=\textwidth]{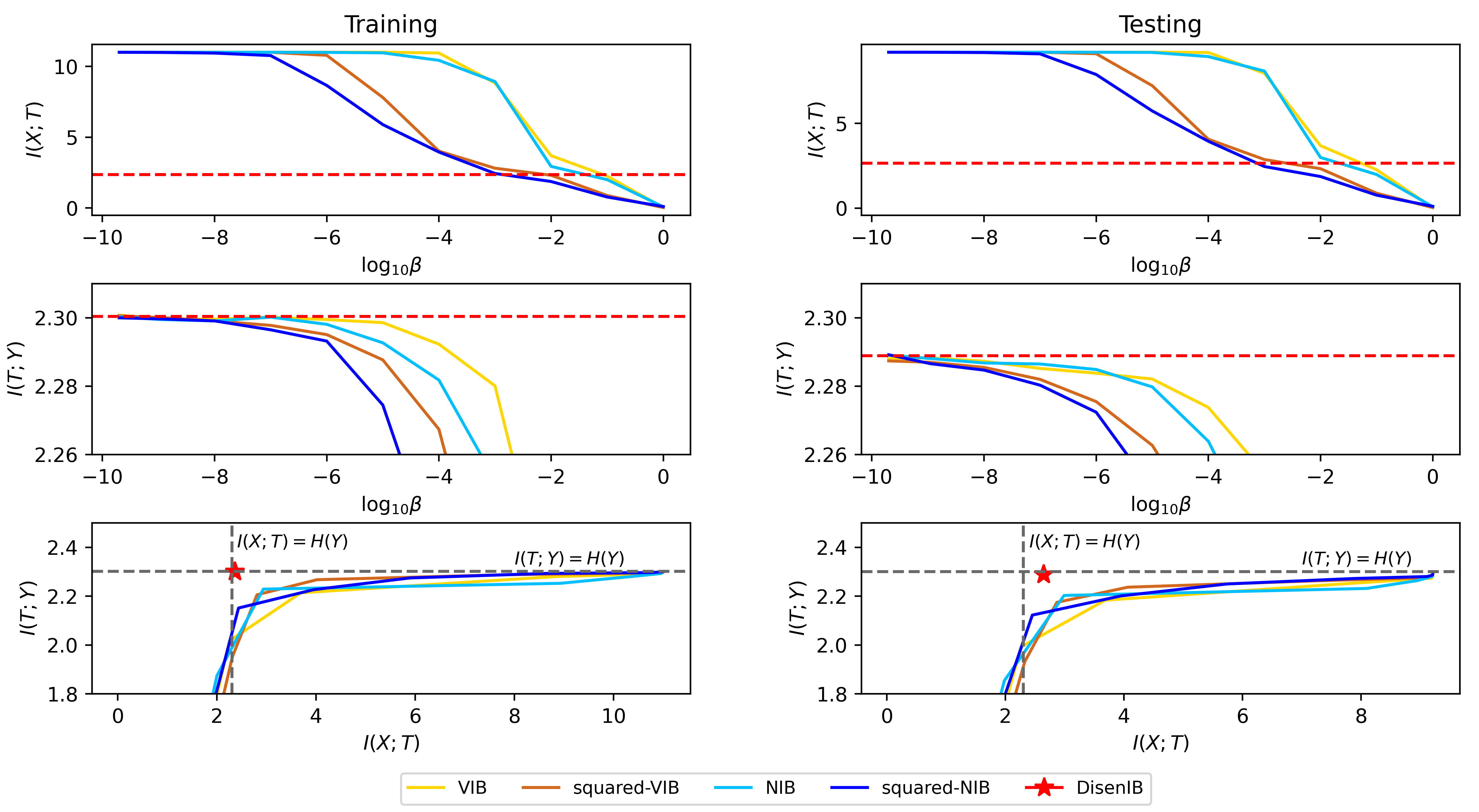}
  \caption{
    MNIST training (\textbf{left}) and testing (\textbf{right}) results. Our method only needs a single optimization to obtain $I\left(X;T\right)$ and $I\left(T;Y\right)$, so our method is represented by red doted line in the top and middle rows, and red star in the bottom row.
    \textbf{Top row:} $I\left(X;T\right)$ \emph{vs.} $\beta$ curve.
    \textbf{Middle row:} $I\left(T;Y\right)$ \emph{vs.} $\beta$ curve.
    \textbf{Bottom row:} IB-plane diagrams defined by $I\left(T;Y\right)$ \emph{vs.} $I\left(X;T\right)$.
  }\label{fig_ib_plane}
\end{figure*}

In this section, we compare existing variants of IB with our method in terms of the behavior on \emph{IB Plane}, which is defined by the axes $I\left(X;T\right)$ (the $x$-axis) and $I\left(T;Y\right)$ (the $y$-axis), showing that our method can perform maximum compression. We report results on both training set and test set. Since all methods use stochastic encoder that produces Gaussian bottleneck, we use Monte Carlo sampling~\cite{goldfeld2019estimating} to get accurate estimate of the $I\left(X;T\right)$ term. To estimate the $I\left(T;Y\right)=H\left(Y\right)-H\left(Y|T\right)$ term, we approximate the conditional entropy with the cross-entropy loss, which is employed in~\cite{kolchinsky2019nonlinear}. $H\left(Y\right)$ is a known constant specified by the dataset. Experimental results on MNIST are summarized in Figure~\ref{fig_ib_plane}, while the results on FashionMNIST, CIFAR10, Tiny-ImageNet, and SUN-RGBD can be found in supplementary.

From Figure~\ref{fig_ib_plane}, we learn that intensive $\beta$ tuning is required for optimizing IB Lagrangian to obtain desired compression level. Specifically, as by~\cite{kolchinsky2018caveats,rodriguez2020convex}, $\beta$ and the compression level are not causally related, and therefore multiple optimizations for different $\beta$ value are required to reach a specific compression level. We can also observe that as $I\left(X;T\right)$ is compressed to $H\left(Y\right)$, $I\left(T;Y\right)$ is inferior to $H\left(Y\right)$, which makes maximum compression (\ie, $I\left(X;T\right)=I\left(T;Y\right)=H\left(Y\right)$) impossible via minimizing the IB Lagrangian. Meanwhile our method can effectively compress $X$ while maintaining $I\left(T;Y\right)$ at a high value. For clarity, we report numerical results at compression level $I\left(X;T\right)=H\left(Y\right)$ as in Table~\ref{table_ib}. From Table~\ref{table_ib}, we observe that $I\left(T;Y\right)$ obtained using our method is barely even reduced and much closer to $H\left(Y\right)$, while $I\left(T;Y\right)$ obtained using other methods are consistently inferior.

\begin{table}[htbp]
	\centering
	\begin{tabular}{cccc}
        \toprule
        \textbf{Dataset}                        & Training              & Testing       & $\beta$      \\
        \midrule
        $H\left(Y\right)$                       & 2.30                  & 2.30          &              \\
        VIB                                     & 2.01                  & 1.99          & $10^{-2}$    \\
        squared-VIB                             & 1.94                  & 1.95          & $10^{-3}$    \\
        NIB                                     & 1.95                  & 1.97          & $10^{-2}$    \\
        squared-NIB                             & 1.98                  & 1.99          & $10^{-3}$    \\
        DisenIB                                 & \textbf{2.25}         & \textbf{2.17} & N/A          \\
        \bottomrule
    \end{tabular}
    \caption{
        $I\left(T;Y\right)$ and $\beta$ value obtained by different methods at compression level $I\left(X;T\right)=H\left(Y\right)$ on MNIST training set and test set. $H\left(Y\right)$ is the amount of $I\left(T;Y\right)$ in the case of maximum compression (Eq.~\eqref{eq_maximum_cprs}).
    }
    \label{table_ib}
\end{table}

For the comparison in terms of generalization, robustness to adversarial attack and out-of-distribution data detection in Sections~\ref{sec_generalize_robust}-\ref{sec_out_of_distribution}, since our results are at the compression level $I\left(X;T\right)=H\left(Y\right)$, we also report results of existing IB variants at the same compression level by tuning $\beta$.

\begin{table*}[h]
	\centering
	\begin{tabular}{cccc}
        \toprule
        \multicolumn{2}{c}{}                &   Training                               & Testing                                    \\
        \midrule
                        &                   &   \multicolumn{2}{c}{
                                                    VIB\ /\
                                                    squared-VIB\ /\
                                                    NIB\ /\
                                                    squared-NIB\ /\
                                                    DisenIB}      \\
        \cline{3-4}
        \multicolumn{2}{c}{Generalization}  &   N/A                                    & 97.6\ /\ 96.2\ /\ 97.2\ /\ 93.3\ /\ \textbf{98.2}     \\
        \midrule
        \multirowcell{3}{
            Adversary \\
            Robustness
        }              &   $\epsilon=0.1$   &   74.1\ /\ 42.1\ /\ 75.2\ /\ 61.3\ /\ \textbf{94.3}    & 73.4\ /\ 42.7\ /\ 75.2\ /\ 62.0\ /\ \textbf{90.2}     \\
                       &   $\epsilon=0.2$   &   19.1\ /\ 8.7\  /\ 21.8\ /\ 24.1\ /\ \textbf{81.5}    & 20.8\ /\ 9.2\  /\ 23.6\ /\ 24.5\ /\ \textbf{80.0}     \\
                       &   $\epsilon=0.3$   &   3.5\  /\ 5.9\  /\ 3.2\  /\ 9.3\  /\ \textbf{68.4}    & 4.2\  /\ 5.9\  /\ 3.4\  /\ 9.9\  /\ \textbf{67.8}     \\
        \bottomrule
    \end{tabular}
    \caption{
        Generalization and adversarial robustness performance (\%) on MNIST dataset.
    }
    \label{table_generalization_attack}
\end{table*}
\begin{table*}[t]
	\centering
	\begin{tabular}{ccc}
        \toprule
        Metric              &   SUN-RGBD~\cite{song2015sun}            & Gaussian Noise                          \\
        \midrule
                            &   \multicolumn{2}{c}{
                                    VIB\ /\
                                    squared-VIB\ /\
                                    NIB\ /\
                                    squared-NIB\ /\
                                    DisenIB}      \\
        \cline{2-3}
        FPR ($95$\% TPR)\ $\downarrow$    &   27.4\ /\ 49.9\ /\ 34.4\ /\ 47.5\ /\ \textbf{0.0}    & 4.5\  /\ 12.7\ /\ 13.4\ /\ 5.3\  /\ \textbf{0.0}     \\
        AUROC\ $\uparrow$                 &   94.6\ /\ 86.6\ /\ 94.2\ /\ 85.6\ /\ \textbf{99.4}   & 98.8\ /\ 95.5\ /\ 97.4\ /\ 90.8\ /\ \textbf{99.7}    \\
        AUPR In\ $\uparrow$               &   94.8\ /\ 83.5\ /\ 95.2\ /\ 83.3\ /\ \textbf{99.6}   & 99.8\ /\ 96.6\ /\ 97.8\ /\ 92.4\ /\ \textbf{99.8}    \\
        AUPR Out\ $\uparrow$              &   93.7\ /\ 83.2\ /\ 91.8\ /\ 83.1\ /\ \textbf{98.9}   & 98.5\ /\ 95.5\ /\ 96.8\ /\ 88.8\ /\ \textbf{99.5}    \\
        Detection Error\ $\downarrow$     &   11.5\ /\ 20.0\ /\ 11.9\ /\ 15.0\ /\ \textbf{1.7}    & 4.7\  /\ 4.7\  /\ 7.6\  /\ 15.7\ /\ \textbf{1.0}     \\
        \bottomrule
    \end{tabular}
    \caption{
        Distinguishing in- and out-of-distribution test data for MNIST image classification (\%). $\uparrow$ (\emph{resp.}, $\downarrow$) indicates that larger (\emph{resp.}, lower) value is better.
    }
    \label{table_out_detection}
\end{table*}

\subsection{Generalization and Adversarial Robustness}\label{sec_generalize_robust}

In this section, we compare our method with existing methods in terms of generalization performance and robustness to adversarial attack. We report the results on MNIST dataset, leaving those on FashionMNIST, CIFAR10, Tiny-ImageNet, and SUN-RGBD in supplementary.

We firstly introduce how to perform evaluation in terms of generalization and robustness to adversarial attack. Generalization performance in Table~\ref{table_generalization_attack} is evaluated by the classification mean accuracy on MNIST test set after training the model on MNIST training set~\cite{alemi2016deep,kolchinsky2019nonlinear}. Considering that deep neural networks can be easily ``fooled'' into making mistakes by changing their inputs by imperceptibly small amounts~\cite{szegedy2013intriguing,goodfellow2014explaining}, adversarial robustness exams how robust the model is to such adversarial examples. To exam adversarial robustness, we use the standard baseline attack method~\cite{goodfellow2014explaining}. Specifically, after training the model on MNIST training set, the training (\emph{resp.}, test) adversary robustness in Table~\ref{table_generalization_attack} are measured by the classification mean accuracy on the adversarial examples of the training (\emph{resp.}, test) set, where the adversarial examples are generated by taking a single step in the gradient direction. We vary $\epsilon\in\left\{0.1,0.2,0.3\right\}$, which controls the magnitude of the perturbation at each pixel~\cite{goodfellow2014explaining}.

We see that our method can slightly outperform existing methods in terms of generalization performance. For adversary robustness, our method is significantly better than existing methods. Compared with our method, existing methods can be easily fooled by making perturbations, which is because a model with degenerated prediction performance (\ie, $I\left(T;Y\right)$) will reduce its adversary robustness~\cite{alemi2016deep}. However, since our method avoids information reduction while compressing, it is more robust to adversarial examples than existing methods.

\subsection{Out-of-distribution Detection}\label{sec_out_of_distribution}

Modern neural networks are known to generalize well when the training and test data are sampled from the same distribution~\cite{zhang2016understanding}. However, when deploying neural networks in real-world applications, there is often very little control over the test data distribution. It is important for classifiers to be aware of uncertainty when shown new types of inputs, \ie, \emph{out-of-distribution} examples. We report experimental results on MNIST dataset, leaving results on FashionMNIST, CIFAR10, Tiny-ImageNet, and SUN-RGBD in supplementary.

We make use of~\cite{liang2017enhancing} for out-of-distribution data detection without re-training the model. To perform detection, the model is first trained on MNIST training set. At detection time, examples from MNIST test set can be viewed as in-distribution data, and we use examples from scene-centric natural image dataset SUN-RGBD~\cite{song2015sun} as out-distribution data, considering the large distribution gap between the two datasets. Following~\cite{liang2017enhancing}, we also use synthetic Gaussian noise dataset which consists of 10,000 random Gaussian noise samples from $\mathcal{N}\left(0, 1\right)$ as out-of-distribution dataset. A good model is expected to accurately detect whether the given example is an out-distribution data or not. In terms of metrics, we use \emph{FPR at 95\% TPR}, \emph{Detection Error}, \emph{AUROC}, \emph{AUPR} following \cite{liang2017enhancing,hendrycks2016baseline}.

\begin{figure*}[h]
    \centering
    \subfigure[MNIST]{\includegraphics[width=0.315\textwidth,height=0.316\textwidth]{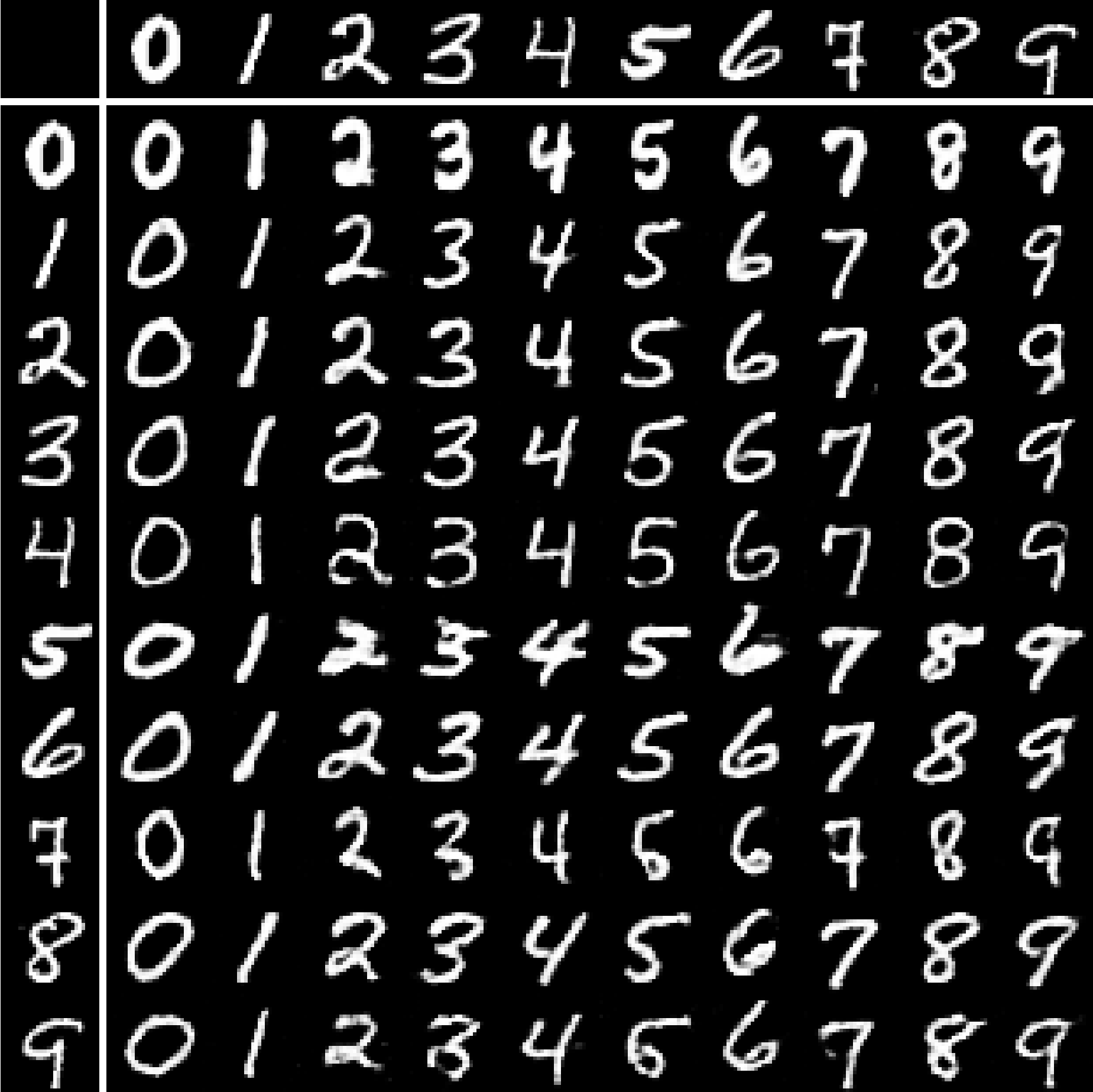}}
    \subfigure[Sprites]{\includegraphics[width=0.3155\textwidth]{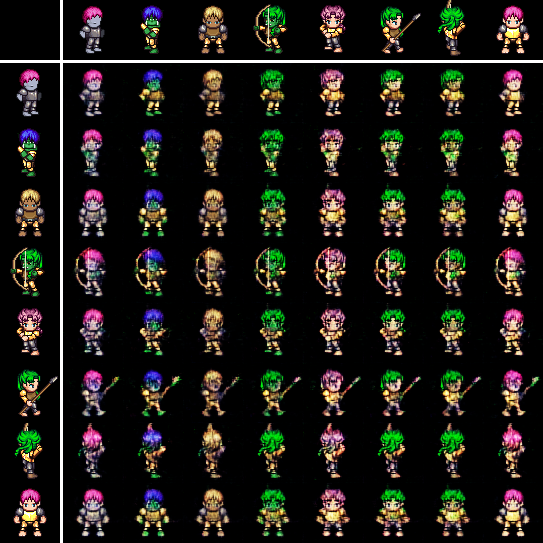}}
    \subfigure[dSprites]{\includegraphics[width=0.32\textwidth,height=0.316\textwidth]{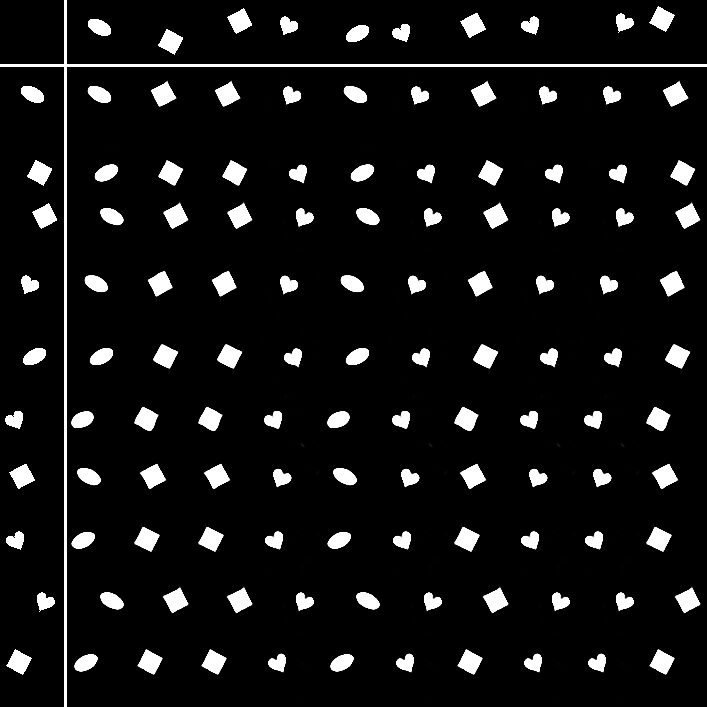}}
    \caption{Visualization grids of image swapping generation. The top row and leftmost column images come from the dataset. The other images are generated using $S$ from leftmost image and $T$ from the image at the top of the column. The diagonal images show reconstructions. }\label{fig_disentangle}
\end{figure*}
\begin{table*}[h]
	\centering
	\begin{tabular}{ccccc}
        \toprule
        \multirowcell{3}{Datasets} & \multicolumn{2}{c}{Classification Accuracy}    & \multicolumn{2}{c}{Information Amount}          \\
                                   &          $S$          &         $T$            &  $I\left(X;T\right)$   &   $I\left(X;S\right)$  \\
        \cline{2-5}
                                   & \multicolumn{4}{c}{
                                        \cite{mathieu2016disentangling}\ /\
                                        \cite{hadad2018two}\ /\
                                        DisenIB}                                                                                      \\
        \midrule
        Training                   &  10.0\ /\ 10.1\ /\ 10.0  &  99.2\ /\ 99.7\ /\ 99.6   &   7.92\ /\ 9.86\ /\ \textbf{2.32}  &  10.75\ /\ 10.81\ /\ 10.86   \\
        \midrule
        Testing                    &  9.9\ /\ 10.1\ /\ 10.1   &  98.2\ /\ 98.3\ /\ 98.2   &   9.60\ /\ 10.17\ /\ \textbf{2.71}  &  9.12\ /\ 9.46\ /\ 9.15   \\
        \bottomrule
    \end{tabular}
    \caption{
        MNIST classification mean accuracy based on $S$ or $T$ (\%) as well as $I\left(X;T\right)$ and $I\left(X;S\right)$.
    }
    \label{table_disentangling}
\end{table*}

The results are summarized in Table~\ref{table_out_detection}. We achieve superior performance than all baselines. For SUN-RGBD~\cite{song2015sun} dataset, our method consistently outperforms the baselines by a large margin, showing that our method is sensitively aware of outliers when given new samples from considerably different data distributions. Since results of existing methods are obtained at the same compression level $I\left(X;T\right)=H\left(Y\right)$ as ours, the prediction performances (\ie, $I\left(T;Y\right)$) are reduced due to the aforementioned trade-off. Because a reduced prediction performance will degenerate a model's capacity of detecting out-of-distribution data~\cite{zhang2016understanding}, detection performances of existing methods are inferior to ours due to the prediction performance reduction.

\subsection{Supervised Disentangling}

We briefly study the disentangling behavior of our method by showing both qualitative and quantitative results. We visualize disentanglement by reconstructing images after \emph{swapping}~\cite{mathieu2016disentangling} data aspects ($S$ or $T$) of different images, which is commonly used for qualitative evaluation in disentangling literature~\cite{mathieu2016disentangling,hadad2018two,higgins2017beta,kim2018disentangling}. In a good disentanglement, given $S$ from one image $I_1$ and $T$ from another image $I_2$, the image generated based on $S$ and $T$ should preserve the $S$-qualities of $I_1$ and the $T$-qualities of $I_2$.
We visualize qualitative results via swapping data aspects on disentanglement benchmark datasets: MNIST, Sprites~\cite{reed2015deep}, and dSprites~\cite{matthey2017dsprites}, where $Y$ represents the digit category, body color, and shape category, respectively. $T$ is $Y$-relevant data aspect and $S$ is $Y$-irrelevant data aspect (\emph{e.g.}, thickness on MNIST, body type on Sprites, and orientation/positision on dSprites).
From Figure~\ref{fig_disentangle}, we learn that our method can separate data aspects $S$ and $T$ well, generating reasonable results when combining data aspects from two different images.

In terms of quantitative metrics, we perform \emph{classification}~\cite{mathieu2016disentangling,hadad2018two} based on both data aspects. To do so, we first train a classifier to predict $Y$ labels based on $S$ or $T$ following~\cite{mathieu2016disentangling,hadad2018two}. For a good disentanglement, high accuracy (\emph{resp.}, random results) should be obtained when applying the classifier on $T$ (\emph{resp.}, $S$). Moreover, to quantify the amount of information of $X$ preserved in each data aspect, we estimate $I\left(X;T\right)$ and $I\left(X;S\right)$ terms.
The quantitative evaluation results are summarized in Table~\ref{table_disentangling}.
We see that the classification performance based on $S$ or $T$ are almost the same across all methods. However, the information controlling behaviors are quite different. While baseline methods leak too much information to the $T$ aspect, our method can exactly control the amount of information stored in the respective aspects.

\section{Conclusion}

In this paper, to perform maximum compression without the need of multiple optimizations, we have implemented the IB method from the perspective of supervised disentangling, introducing the Disentangled Information Bottleneck (DisenIB). Theoretical and experimental results have demonstrated that our method is consistent on maximum compression, and performs well in terms of generalization, robustness to adversarial attack, out-of-distribution data detection, and supervised disentangling.

\section{Acknowledgement}
The work is supported by the National Key R\&D Program of China (2018AAA0100704) and the Shanghai Science and Technology R\&D Program of China (20511100300) and is partially sponsored by National Natural Science Foundation of China (Grant No.61902247) and Shanghai Sailing Program (19YF1424400).

\bibliography{references}



\section*{Supplementary material}

# Supplementary for Disentangled Information Bottleneck

In Section 1, we prove Theorem 1-2 in the main text. In Section 2 we provide practical implementation details. Experimental results on more datasets can be found in Section 3.

## 1. Proofs

In this section, we prove Theorem 1 that analyzes the IB Lagrangian trade-off, and Theorem 2 that shows the consistency property of our proposed DisenIB.

### 1.1. Proof of Theorem 1

Recall the Theorem 1 given in the main text:

**Theorem 1.** *Consider the derivable IB Lagrangian,*

$$\mathcal{L}_{\mathrm{IB}}\left[q\left(T|X\right);\beta\right] = -I\left(T;Y\right) + \beta I\left(X;T\right), \tag{1}$$

*to be minimized over $q$ with $\beta \geqslant 0$. Let $q_\beta^*$ optimize $\mathcal{L}_{\mathrm{IB}}\left[q\left(T|X\right);\beta\right]$. Assume that $I_{q_\beta^*}\left(X;T\right) \neq 0$,*

$$\frac{\partial I_{q_\beta^*}\left(T;Y\right)}{\partial \beta} < 0 \text{ and } \frac{\partial I_{q_\beta^*}\left(X;T\right)}{\partial \beta} < 0. \tag{2}$$

*Proof.* From Lemma 1 (see Section 1.3), we learn that

$$\frac{\partial I_{q_\beta^*}\left(T;Y\right)}{\partial \beta} \leqslant 0 \text{ and } \frac{\partial I_{q_\beta^*}\left(X;T\right)}{\partial \beta} \leqslant 0 \tag{3}$$

by treating $-I\left(T;Y\right)$, $I\left(X;T\right)$ and $q$ as $A\left(\kappa\right)$, $B\left(\kappa\right)$ and $\kappa$ respectively. Moreover, we also learn that to prove that both equalities in Eq. (3) can not hold is equivalent to proving that[1]

$$\exists \rho, \left\langle \left.\frac{\delta I\left(X;T\right)}{\delta q}\right|_{q=q_\beta^*}, \rho \right\rangle \neq 0. \tag{4}$$

By leveraging Lemma 2 (see Section 1.3), we learn that $\forall q, \left\langle \frac{\delta I(X;T)}{\delta q}, \hat{q} - q \right\rangle < 0$, where $\hat{q}$ denoting the aggregated distribution $q\left(T\right)$. Due to the arbitrariness of $q$,

$$\left\langle \frac{\delta I\left(X;T\right)}{\delta q}, \hat{q} - q \right\rangle \bigg|_{q=q_\beta^*} = \left\langle \left.\frac{\delta I\left(X;T\right)}{\delta q}\right|_{q=q_\beta^*}, \left.\hat{q} - q\right|_{q=q_\beta^*} \right\rangle < 0, \tag{5}$$

Hence $\exists \rho = \left.\hat{q} - q\right|_{q=q_\beta^*}$, such that $\left\langle \left.\frac{\delta I(X;T)}{\delta q}\right|_{q=q_\beta^*}, \rho \right\rangle \neq 0$. Therefore by leveraging Lemma 1, both equalities in Eq. (3) can not hold, *i.e.*,

$$\frac{\partial I_{q_\beta^*}\left(T;Y\right)}{\partial \beta} < 0 \text{ and } \frac{\partial I_{q_\beta^*}\left(X;T\right)}{\partial \beta} < 0, \tag{6}$$

and Theorem 1 is proved. □

---
[1]Functional derivatives [1] are used, since $I\left(X;T\right)$ is a functional of probabilistic mapping $q$.

## 1.2. Proof of Theorem 2

Recall the DisenIB objective functional to be minimized in the main text:

$$\mathcal{L}_{\text{DisenIB}}\left[q\left(S|X\right),q\left(T|X\right)\right] = -I\left(T;Y\right) - I\left(X;S,Y\right) + I\left(S;T\right), \tag{7}$$

and the definition of the *consistency* property on maximum compression as well as Theorem 2:

**Definition 1** (Consistency). *The lower-bounded cost functional $\mathcal{L}$ is consistent on maximum compression, if*

$$\forall \epsilon > 0, \exists \delta > 0, \quad \mathcal{L} - \mathcal{L}^* < \delta \rightarrow |I\left(X;T\right) - H\left(Y\right)| + |I\left(T;Y\right) - H\left(Y\right)| < \epsilon, \tag{8}$$

*where $\mathcal{L}^*$ is the global minimum of $\mathcal{L}$.*

**Theorem 2.** $\mathcal{L}_{\text{DisenIB}}$ *is consistent on maximum compression.*

*Proof.* The global minimum of $\mathcal{L}_{\text{DisenIB}}$ is

$$\mathcal{L}^*_{\text{DisenIB}} = \min_{q(T|X), q(S|X)} \mathcal{L}_{\text{DisenIB}} \tag{9}$$

$$= -\max_{q(T|X)} I\left(T;Y\right) - \max_{q(S|X)} I\left(X;S,Y\right) + \min_{q(T|X), q(S|X)} I\left(S;T\right) \tag{10}$$

$$= -H\left(Y\right) - H\left(X\right). \tag{11}$$

Assume that $\mathcal{L}_{\text{DisenIB}} - \mathcal{L}^*_{\text{DisenIB}} < \delta$, then we have that

$$H\left(Y\right) - I\left(T;Y\right) < \delta \text{ and } H\left(X\right) - I\left(X;S,Y\right) < \delta \text{ and } I\left(S;T\right) < \delta. \tag{12}$$

Given Markov chains $S \leftrightarrow X \leftrightarrow Y$, $S \leftrightarrow X \leftrightarrow T$ and $T \leftrightarrow X \leftrightarrow Y$, by leveraging Lemma 3 (see Section 1.3), we have that

$$I\left(X;S,Y\right) + I\left(S;Y\right) = I\left(X;S\right) + I\left(X;Y\right), \tag{13}$$

$$I\left(X;S,T\right) + I\left(S;T\right) = I\left(X;S\right) + I\left(X;T\right), \tag{14}$$

$$I\left(X;T,Y\right) + I\left(T;Y\right) = I\left(X;T\right) + I\left(X;Y\right). \tag{15}$$

By subtracting Eq. (14) from Eq. (13), we can derive Eq. (16). Then by leveraging inequality in Eq. (12), we can derive Eq. (17):

$$I\left(X;T\right) + I\left(S;Y\right) - I\left(X;Y\right) - I\left(X;S,T\right) = -I\left(X;S,Y\right) + I\left(S;T\right) \tag{16}$$

$$< 2\delta - H\left(X\right). \tag{17}$$

Since $I\left(X;Y\right) = H\left(Y\right) - H\left(Y|X\right) = H\left(Y\right)$, by rearranging terms in Eq. (15) and leveraging inequalities in Eq. (12), we have that

$$I\left(X;T,Y\right) - I\left(X;T\right) = I\left(X;Y\right) - I\left(T;Y\right) = H\left(Y\right) - I\left(T;Y\right) < \delta. \tag{18}$$

By substituting $I\left(X;T\right)$ in Eq. (16) for $I\left(X;T,Y\right)$ and leveraging inequality in Eq. (17)-(18),

$$I\left(X;T,Y\right) + I\left(S;Y\right) - I\left(X;Y\right) - I\left(X;S,T\right) < 3\delta - H\left(X\right). \tag{19}$$

By moving $H\left(X\right)$ from the right side to the left side, we further have that

$$H\left(X\right) - I\left(X;S,T\right) + I\left(X;T,Y\right) - I\left(X;Y\right) + I\left(S;Y\right) < 3\delta. \tag{20}$$

Since $H\left(X\right) - I\left(X;S,T\right) \geqslant 0$, $I\left(X;T,Y\right) - I\left(X;Y\right) \geqslant 0$ and $I\left(S;Y\right) \geqslant 0$, we have that

$$H\left(X\right) - I\left(X;S,T\right) \leqslant 3\delta \text{ and } I\left(X;T,Y\right) - I\left(X;Y\right) \leqslant 3\delta \text{ and } I\left(S;Y\right) \leqslant 3\delta. \tag{21}$$

By plugging Eq. (21) into Eq. (15), we have that $I\left(X;T\right) - I\left(T;Y\right) \leqslant 3\delta$. By leveraging the *data processing inequality* (DPI) [2], we have that $I\left(X;T\right) - I\left(T;Y\right) \geqslant 0$. Therefore

$$|I\left(X;T\right) - I\left(T;Y\right)| \leqslant 3\delta. \tag{22}$$

By rearranging terms in Eq. (16), and using inequalities in Eq. (12) and Eq. (21), we have that

$$I(X;T) - I(X;Y) = I(X;S,T) + I(S;T) - I(X;S,Y) - I(S;Y) > -6\delta, \tag{23}$$
$$I(X;T) - I(X;Y) = I(X;S,T) + I(S;T) - I(X;S,Y) - I(S;Y) < 2\delta, \tag{24}$$

therefore we further have that $|I(X;T) - I(X;Y)| < 6\delta$. By leveraging Eq. (22), we have that

$$|I(T;Y) - I(X;Y)| \leqslant |I(X;T) - I(T;Y)| + |I(X;T) - I(X;Y)| < 9\delta, \tag{25}$$

therefore we have that

$$|I(X;T) - I(X;Y)| + |I(T;Y) - I(X;Y)| < 15\delta. \tag{26}$$

Since $I(X;Y) = H(Y)$, we have that $\forall \epsilon > 0, \exists \delta = \frac{\epsilon}{15} > 0$, such that

$$\mathcal{L}_{\text{DisenIB}} - \mathcal{L}^*_{\text{DisenIB}} < \delta \to |I(X;T) - H(Y)| + |I(T;Y) - H(Y)| < \epsilon, \tag{27}$$

therefore $\mathcal{L}_{\text{DisenIB}}$ is consistent on maximum compression. □

### 1.3. Auxiliary Lemmas

We provide the following Lemmas that are used in the analyses above.

**Lemma 1** ([3], Proposition 4). *Consider a derivable functional given by a sum of two objectives,*

$$O(\kappa; \beta) = A(\kappa) + \beta B(\kappa), \tag{28}$$

*to be minimized over parameter $\kappa$, and $\beta \geqslant 0$ is a hyper-parameter. Let $\kappa^*_\beta$ optimize $O(\kappa; \beta)$,*

$$\frac{\partial A\left(\kappa^*_\beta\right)}{\partial \beta} \geqslant 0 \text{ and } \frac{\partial B\left(\kappa^*_\beta\right)}{\partial \beta} \leqslant 0, \tag{29}$$

*where both equalities hold $\iff \forall \rho, \left\langle \frac{\delta B}{\delta \kappa}\big|_{\kappa=\kappa^*_\beta}, \rho \right\rangle = 0$.*

*Proof.* We first prove Eq. (29). Let $\beta_2 > \beta_1$,

$$O\left(\kappa^*_{\beta_2}; \beta_1\right) - O\left(\kappa^*_{\beta_1}; \beta_1\right) \tag{30}$$
$$= O\left(\kappa^*_{\beta_2}; \beta_1\right) - O\left(\kappa^*_{\beta_2}; \beta_2\right) + O\left(\kappa^*_{\beta_2}; \beta_2\right) - O\left(\kappa^*_{\beta_1}; \beta_2\right) + O\left(\kappa^*_{\beta_1}; \beta_2\right) - O\left(\kappa^*_{\beta_1}; \beta_1\right) \tag{31}$$
$$= (\beta_1 - \beta_2) B\left(\kappa^*_{\beta_2}\right) + O\left(\kappa^*_{\beta_2}; \beta_2\right) - O\left(\kappa^*_{\beta_1}; \beta_2\right) + (\beta_2 - \beta_1) B\left(\kappa^*_{\beta_1}\right) \tag{32}$$
$$= (\beta_2 - \beta_1) \left[B\left(\kappa^*_{\beta_1}\right) - B\left(\kappa^*_{\beta_2}\right)\right] + O\left(\kappa^*_{\beta_2}; \beta_2\right) - O\left(\kappa^*_{\beta_1}; \beta_2\right), \tag{33}$$

and we further derive that

$$(\beta_2 - \beta_1) \left[B\left(\kappa^*_{\beta_1}\right) - B\left(\kappa^*_{\beta_2}\right)\right] = \underbrace{\left[O\left(\kappa^*_{\beta_2}; \beta_1\right) - O\left(\kappa^*_{\beta_1}; \beta_1\right)\right]}_{\geqslant 0} - \underbrace{\left[O\left(\kappa^*_{\beta_2}; \beta_2\right) - O\left(\kappa^*_{\beta_1}; \beta_2\right)\right]}_{\leqslant 0} \tag{34}$$
$$\geqslant 0,$$

where $\left[O\left(\kappa^*_{\beta_2}; \beta_1\right) - O\left(\kappa^*_{\beta_1}; \beta_1\right)\right] \geqslant 0$ and $\left[O\left(\kappa^*_{\beta_2}; \beta_2\right) - O\left(\kappa^*_{\beta_1}; \beta_2\right)\right] \leqslant 0$ follow from the optimality of $\kappa^*_{\beta_1}$ at $\beta = \beta_1$ and $\kappa^*_{\beta_2}$ at $\beta = \beta_2$. Therefore, $\beta_2 > \beta_1 \implies B\left(\kappa^*_{\beta_2}\right) \leqslant B\left(\kappa^*_{\beta_1}\right)$. We then prove $\beta_2 > \beta_1 \implies A\left(\kappa^*_{\beta_2}\right) \geqslant A\left(\kappa^*_{\beta_1}\right)$ using contradiction. Assume that $A\left(\kappa^*_{\beta_2}\right) < A\left(\kappa^*_{\beta_1}\right)$, then

$$O\left(\kappa^*_{\beta_2}; \beta_1\right) = A\left(\kappa^*_{\beta_2}\right) + \beta_1 B\left(\kappa^*_{\beta_2}\right) < A\left(\kappa^*_{\beta_1}\right) + \beta_1 B\left(\kappa^*_{\beta_1}\right) = O\left(\kappa^*_{\beta_1}; \beta_1\right), \tag{35}$$

which contradicts the optimality of $\kappa^*_{\beta_1}$ at $\beta = \beta_1$. Therefore, $\beta_2 > \beta_1 \implies A\left(\kappa^*_{\beta_2}\right) \geqslant A\left(\kappa^*_{\beta_1}\right)$. Hence Eq. (29) is proved. We then prove that

$$\frac{\partial A\left(\kappa^*_\beta\right)}{\partial \beta} = \frac{\partial B\left(\kappa^*_\beta\right)}{\partial \beta} = 0 \iff \forall \rho, \left\langle \frac{\delta B}{\delta \kappa}\bigg|_{\kappa=\kappa^*_\beta}, \rho \right\rangle = 0. \tag{36}$$

**Sufficiency** Since $\frac{\partial A(\kappa_\beta^*)}{\partial \beta} = \frac{\partial B(\kappa_\beta^*)}{\partial \beta} = 0$, we have that

$$\lim_{\beta' \to \beta} \frac{O(\kappa_{\beta'}^*; \beta') - O(\kappa_\beta^*; \beta')}{\beta' - \beta} = \lim_{\beta' \to \beta} \frac{A(\kappa_{\beta'}^*) - A(\kappa_\beta^*)}{\beta' - \beta} + \beta' \lim_{\beta' \to \beta} \frac{B(\kappa_{\beta'}^*) - B(\kappa_\beta^*)}{\beta' - \beta} = 0. \quad (37)$$

We then prove sufficiency using contradiction. Assume that $\exists \rho, \left\langle \frac{\delta B}{\delta \kappa} \big|_{\kappa = \kappa_\beta^*}, \rho \right\rangle \neq 0$. For concise mathematical expression, we use $O^\beta \triangleq A + \beta B$. Since

$$\left\langle \frac{\delta O^\beta}{\delta \kappa}, \rho \right\rangle = \left\langle \frac{\delta A}{\delta \kappa}, \rho \right\rangle + \beta \left\langle \frac{\delta B}{\delta \kappa}, \rho \right\rangle, \quad (38)$$

given $\beta_2 = \beta_1 + \Delta\beta$ with $\Delta\beta > 0$, we have that

$$\left\langle \frac{\delta O^{\beta_2}}{\delta \kappa} \bigg|_{\kappa = \kappa_{\beta_1}^*}, \rho \right\rangle - \left\langle \frac{\delta O^{\beta_1}}{\delta \kappa} \bigg|_{\kappa = \kappa_{\beta_1}^*}, \rho \right\rangle = \Delta\beta \left\langle \frac{\delta B}{\delta \kappa} \bigg|_{\kappa = \kappa_{\beta_1}^*}, \rho \right\rangle. \quad (39)$$

Since $\left\langle \frac{\delta O^{\beta_1}}{\delta \kappa} \big|_{\kappa = \kappa_{\beta_1}^*}, \rho \right\rangle = 0$ due to the optimality of $\kappa_{\beta_1}^*$, we have that

$$\left\langle \frac{\delta O^{\beta_2}}{\delta \kappa} \bigg|_{\kappa = \kappa_{\beta_1}^*}, \rho \right\rangle = \Delta\beta \left\langle \frac{\delta B}{\delta \kappa} \bigg|_{\kappa = \kappa_{\beta_1}^*}, \rho \right\rangle \neq 0. \quad (40)$$

Since $\left\langle \frac{\delta O^{\beta_2}}{\delta \kappa} \big|_{\kappa = \kappa_{\beta_1}^*}, \rho \right\rangle = \frac{d}{d\epsilon} O^{\beta_2}\left(\kappa_{\beta_1}^* + \epsilon\rho\right) \big|_{\epsilon = 0} \neq 0$, the following holds:

$$\exists \rho', \text{ such that } \lim_{\epsilon \to 0} \frac{O^{\beta_2}\left(\kappa_{\beta_1}^* + \epsilon\rho'\right) - O^{\beta_2}\left(\kappa_{\beta_1}^*\right)}{\epsilon} < 0. \quad (41)$$

Therefore, we further have that

$$\lim_{\epsilon \to 0} \frac{1}{\epsilon} \left( \frac{O^{\beta_2}\left(\kappa_{\beta_1}^* + \epsilon\rho'\right) - O^{\beta_2}\left(\kappa_{\beta_2}^*\right)}{\Delta\beta} + \frac{O^{\beta_2}\left(\kappa_{\beta_2}^*\right) - O^{\beta_2}\left(\kappa_{\beta_1}^*\right)}{\Delta\beta} \right) < 0. \quad (42)$$

By leveraging Eq. (37) and pushing $\Delta\beta \to 0$, we further have that

$$\lim_{\Delta\beta \to 0} \frac{1}{\Delta\beta} \left( \lim_{\epsilon \to 0} \frac{O^{\beta_2}\left(\kappa_{\beta_1}^* + \epsilon\rho'\right) - O^{\beta_2}\left(\kappa_{\beta_2}^*\right)}{\epsilon} \right) < 0, \quad (43)$$

namely $\lim_{\epsilon \to 0} \frac{O^{\beta_2}(\kappa_{\beta_1}^* + \epsilon\rho') - O^{\beta_2}(\kappa_{\beta_2}^*)}{\epsilon} < 0$, which contradicts the optimality of $\kappa_{\beta_2}^*$ at $\beta = \beta_2$. Therefore we have that

$$\frac{\partial A(\kappa_\beta^*)}{\partial \beta} = \frac{\partial B(\kappa_\beta^*)}{\partial \beta} = 0 \implies \forall \rho, \left\langle \frac{\delta B}{\delta \kappa} \bigg|_{\kappa = \kappa_\beta^*}, \rho \right\rangle = 0. \quad (44)$$

**Necessity** Since $\forall \rho, \left\langle \frac{\delta B}{\delta \kappa} \big|_{\kappa = \kappa_\beta^*}, \rho \right\rangle = 0$, from Eq. (46), we have that $\frac{\partial B(\kappa_\beta^*)}{\partial \beta} = 0$. We then show that $\frac{\partial A(\kappa_\beta^*)}{\partial \beta}$ and $\frac{\partial B(\kappa_\beta^*)}{\partial \beta}$ have the same sign and thus $\frac{\partial A(\kappa_\beta^*)}{\partial \beta} = \frac{\partial B(\kappa_\beta^*)}{\partial \beta} = 0$. Obviously, following the optimality of $\kappa_\beta^*$, we have that

$$\frac{\delta A}{\delta \kappa} \bigg|_{\kappa = \kappa_\beta^*} = \beta \frac{\delta B}{\delta \kappa} \bigg|_{\kappa = \kappa_\beta^*}. \quad (45)$$

Since that

$$\frac{\partial A\left(\kappa_\beta^*\right)}{\partial \beta} = \left\langle \frac{\delta A}{\delta \kappa}\bigg|_{\kappa=\kappa_\beta^*}, \frac{\partial \kappa_\beta^*}{\partial \beta}\right\rangle \quad \text{and} \quad \frac{\partial B\left(\kappa_\beta^*\right)}{\partial \beta} = \left\langle \frac{\delta B}{\delta \kappa}\bigg|_{\kappa=\kappa_\beta^*}, \frac{\partial \kappa_\beta^*}{\partial \beta}\right\rangle, \tag{46}$$

by plugging Eq. (45) into Eq. (46), we further have that

$$\frac{\partial A\left(\kappa_\beta^*\right)}{\partial \beta} = \beta \frac{\partial B\left(\kappa_\beta^*\right)}{\partial \beta}, \tag{47}$$

which shows that $\frac{\partial A(\kappa_\beta^*)}{\partial \beta}$ and $\frac{\partial B(\kappa_\beta^*)}{\partial \beta}$ have the same sign. Therefore the necessity is proved:

$$\frac{\partial A\left(\kappa_\beta^*\right)}{\partial \beta} = \frac{\partial B\left(\kappa_\beta^*\right)}{\partial \beta} = 0 \impliedby \forall \rho, \left\langle \frac{\delta B}{\delta \kappa}\bigg|_{\kappa=\kappa_\beta^*}, \rho\right\rangle = 0. \tag{48}$$

□

**Lemma 2.** *Assume that $I(X;T) \neq 0$. $\forall q$, given a small perturbation $\epsilon(\hat{q} - q)$ on $q$ with $\epsilon$ being a positive infinitesimal, $I_q(X;T)$ strictly decreases. Formally,*

$$\left\langle \frac{\delta I(X;T)}{\delta q}, \hat{q} - q \right\rangle = \frac{\mathrm{d}}{\mathrm{d}\epsilon} I_{q+\epsilon(\hat{q}-q)}(X;T)\bigg|_{\epsilon=0} < 0. \tag{49}$$

*Proof.* We introduce notations for the sake of concise mathematical expression. Assume that associated dataset $\{(x_i, y_i)\}_{i=1}^N$ contains $N$ i.i.d. pairs, *i.e.*, the empirical data distribution $p_{\mathrm{data}}(x_i) = \frac{1}{N}$. Let $F(q)$ denote the entropy under $q$ (*i.e.*, $F(q) \triangleq -\int q(t) \log q(t) \, \mathrm{d}t$), and $I(q)$ denote $I(X;T)$ under $q$ (*i.e.*, $I(q) \triangleq I_q(X;T)$). Let $q_i \triangleq q(T|x_i)$, $i = 1, 2, ..., N$, and $\hat{q} \triangleq \frac{1}{N}\sum_{i=1}^N q_i = q(T)$.

Since $I(q) \neq 0$, the case where $q_1 = q_2 = \cdots = q_N$ can not hold. To prove this, contradiction can be used. Assume that $q_1 = q_2 = \cdots = q_N$. Since $F(q)$ is strictly concave [2], we have that

$$F(\hat{q}) = \frac{1}{N}\sum_{i=1}^N F(q_i). \tag{50}$$

From Eq. (50), we learn that $I(q) = F(\hat{q}) - \frac{1}{N}\sum_{i=1}^N F(q_i) = 0$, which contradicts $I(q) = 0$.

We then prove Eq. (49). We have that

$$\left\langle \frac{\delta I}{\delta q}, \hat{q} - q \right\rangle = \left\langle \frac{\delta\left[F(\hat{q}) - \frac{1}{N}\sum_{i=1}^N F(q_i)\right]}{\delta q}, \hat{q} - q \right\rangle \tag{51}$$

$$= \left\langle \frac{\delta F}{\delta q}\bigg|_{q=\hat{q}}, \hat{q} - q \right\rangle - \frac{1}{N}\sum_{i=1}^N \left\langle \frac{\delta F}{\delta q}\bigg|_{q=q_i}, \hat{q} - q \right\rangle. \tag{52}$$

Since that

$$\left\langle \frac{\delta F}{\delta q}, \hat{q} - q \right\rangle = \frac{\mathrm{d}}{\mathrm{d}\epsilon} F(q + \epsilon(\hat{q} - q))\bigg|_{\epsilon=0} \tag{53}$$

$$= \frac{\mathrm{d}}{\mathrm{d}\epsilon}\left(-\int (q + \epsilon(\hat{q} - q)) \log(q + \epsilon(\hat{q} - q)) \, \mathrm{d}t\right)\bigg|_{\epsilon=0} \tag{54}$$

$$= -\int \frac{\mathrm{d}\left((q + \epsilon(\hat{q} - q))\log(q + \epsilon(\hat{q} - q))\right)}{\mathrm{d}\epsilon} \mathrm{d}t\bigg|_{\epsilon=0} \tag{55}$$

$$= -\int \left(\log(q + \epsilon(\hat{q} - q))(\hat{q} - q) + (\hat{q} - q)\right) \mathrm{d}t\bigg|_{\epsilon=0} \tag{56}$$

$$= -\int (\log q + 1)(\hat{q} - q) \, \mathrm{d}t, \tag{57}$$

we further derive that
$$\left\langle \frac{\delta F}{\delta q}\bigg|_{q=\hat{q}}, \hat{q} - q \right\rangle = 0, \tag{58}$$

and
$$\frac{1}{N}\sum_{i=1}^{N}\left\langle \frac{\delta F}{\delta q}\bigg|_{q=q_i}, \hat{q} - q \right\rangle = -\frac{1}{N}\sum_{i=1}^{N}\int (\log q_i + 1)(\hat{q} - q_i)\,\mathrm{d}t \tag{59}$$

$$= \int \frac{1}{N}\sum_{i=1}^{N}(q_i - \hat{q})\log q_i \,\mathrm{d}t \tag{60}$$

$$= \int \frac{1}{N}\sum_{i=1}^{N}q_i \log q_i \,\mathrm{d}t - \int \hat{q}\frac{1}{N}\sum_{i=1}^{N}\log q_i \,\mathrm{d}t. \tag{61}$$

Since $\log q$ is strictly concave and the case where $q_1 = q_2 = \cdots q_N$ can not hold, we have that
$$\frac{1}{N}\sum_{i=1}^{N}\log q_i < \log \frac{1}{N}\sum_{i=1}^{N} q_i = \log \hat{q}. \tag{62}$$

By plugging Eq. (62) into Eq. (61), we can derive that
$$\frac{1}{N}\sum_{i=1}^{N}\left\langle \frac{\delta F}{\delta q}\bigg|_{q=q_i}, \hat{q} - q \right\rangle > \int \frac{1}{N}\sum_{i=1}^{N}q_i \log q_i \,\mathrm{d}t - \int \hat{q}\log \hat{q}\,\mathrm{d}t. \tag{63}$$

Since $q \log q$ is strictly convex and the case where $q_1 = q_2 = \cdots q_N$ can not hold, following Jessen's inequality [2], we have that
$$\frac{1}{N}\sum_{i=1}^{N} q_i \log q_i > \hat{q}\log \hat{q}. \tag{64}$$

By plugging Eq. (64) into Eq. (63), we further derive that
$$\frac{1}{N}\sum_{i=1}^{N}\left\langle \frac{\delta F}{\delta q}\bigg|_{q=q_i}, \hat{q} - q \right\rangle > 0. \tag{65}$$

By plugging Eq. (58) and Eq. (65) into Eq. (52), we finally derive that
$$\left\langle \frac{\delta I}{\delta q}, \hat{q} - q \right\rangle < 0. \tag{66}$$

□

**Lemma 3.** *Let $X, Y, Z$ be random variables. Assume that the probabilistic mapping $p(X, Y, Z)$ follows from Markov chain $X \leftrightarrow Y \leftrightarrow Z$. Then*
$$I(Y; X, Z) = I(Y; X) + I(Y; Z) - I(X; Z). \tag{67}$$

*Proof.* By leveraging the Markov chain $X \leftrightarrow Y \leftrightarrow Z$,
$$I(Y; X, Z) = \mathbb{E}_{p(x,y,z)} \log \frac{p(x, y, z)}{p(y)\,p(x, z)} \tag{68}$$

$$= \mathbb{E}_{p(x,y,z)} \log \frac{p(y)\,p(x|y)\,p(z|y)}{p(y)\,p(x, z)} \tag{69}$$

$$= \mathbb{E}_{p(x,y,z)} \log \frac{p(x|y)\,p(y)}{p(y)\,p(x)} \cdot \frac{p(z|y)\,p(y)}{p(y)\,p(z)} \cdot \frac{p(x)\,p(z)}{p(x, z)} \tag{70}$$

$$= I(Y; X) + I(Y; Z) - I(X; Z). \tag{71}$$

□

## 2. Implementation Details

In the following analyses, for concise mathematical expression, we omit the conventions that we use. Specifically, we assume that the training dataset $\mathcal{D}$ is given as $\mathcal{D} = \{(x_i, y_i)\}_{i=1}^{N}$, where the sample $x_i \in \mathcal{X}$ is a realization from a random variable $X$, the label $y_i \in \mathcal{Y}$ is a realization from a random variable $Y$, and the pairs $(x_i, y_i)$ are i.i.d. samples from a joint probability distribution $p_{\text{data}}(X, Y)$.

In main text, we introduce variational mappings $p(y|t)$ and $r(x|s, y)$ to lower bound $I(T; Y)$ and $I(X; S, Y)$,

$$I(T; Y) = \mathbb{E}_{q(y,t)} \log q(y|t) - \mathbb{E}_{q(y)} \log q(y) \geqslant \mathbb{E}_{q(y,t)} \log p(y|t) + H(Y), \tag{72}$$

$$I(X; S, Y) = \mathbb{E}_{q(x,s,y)} \log q(x|s, y) - \mathbb{E}_{q(x)} \log q(x) \geqslant \mathbb{E}_{q(x,s,y)} \log r(x|s, y) + H(X). \tag{73}$$

By rewriting (leveraging Markov chains $Y \leftrightarrow X \leftrightarrow T$ and $Y \leftrightarrow X \leftrightarrow S$)

$$q(y, t) = \mathbb{E}_{p_{\text{data}}(x)} p_{\text{data}}(y|x) q(t|x), \tag{74}$$

$$q(x, s, y) = p_{\text{data}}(x) p_{\text{data}}(y|x) q(s|x), \tag{75}$$

and dropping constant $H(X)$ and $H(Y)$, we have that

$$\begin{aligned} I(T; Y) + I(X; S, Y) &\geqslant \mathbb{E}_{q(y,t)} \log p(y|t) + \mathbb{E}_{q(x,s,y)} \log r(x|s, y) \\ &= \mathbb{E}_{p_{\text{data}}(x)} \mathbb{E}_{p_{\text{data}}(y|x)} \left[ \mathbb{E}_{q(t|x)} \log p(y|t) + \mathbb{E}_{q(s|x)} \log r(x|s, y) \right]. \end{aligned} \tag{76}$$

Therefore minimizing $-I(T; Y) - I(X; S, Y)$ is equivalent to

$$\min_{q,p,r} \mathbb{E}_{p_{\text{data}}(x)} \mathbb{E}_{p_{\text{data}}(y|x)} \left[ -\mathbb{E}_{q(t|x)} \log p(y|t) - \mathbb{E}_{q(s|x)} \log r(x|s, y) \right]. \tag{77}$$

We demonstrate how to practically implement DisenIB in the scenario of classification, while the regression case can be similarly obtained. In practice, the probabilistic mappings $q(t|x)$ and $q(s|x)$ are parameterized by two *encoders*: $E_t : \mathcal{X} \to \mathbb{R}^K$ and $E_s : \mathcal{X} \to \mathbb{R}^K$, where $E_t$ and $E_s$ are used to produce bottleneck representation $t$ and $s$ respectively, and $K$ is the dimensionality of bottleneck representation. Since in deterministic scenario (*i.e.*, $t$ is a deterministic function of $x$), the MI $I(X; T)$ is piecewise constant, making the optimization problem ill-posed [4], we make $E_t$ stochastic by injecting Gaussian noise $\mathcal{N}(\mathbf{0}, \sigma^2 \mathbf{I})$, using reparameterization trick [5, 6] to produce $t$,

$$t \sim E_t(x) + \mathcal{N}(\mathbf{0}, \sigma^2 \mathbf{I}). \tag{78}$$

For simplicity, we treat output of $E_t$ as representation obtained after resampling trick. In terms of another bottleneck $S$, we conduct the same process as $T$. The variational approximation $p(y|t)$ is parameterized by a *decoder* $D : \mathbb{R}^K \to \mathbb{R}^{|\mathcal{Y}|}$, which produce probabilities over possible outcomes $y$, Then, one can easily see that $-\mathbb{E}_{p_{\text{data}}(y|x)} \mathbb{E}_{q(t|x)} \log p(y|t)$ is the *Cross Entropy* loss,

$$\mathcal{L}_{\text{CE}}(D(E_t(x)), y) = -\log D(E_t(x))_y, \tag{79}$$

where $D(E_t(x))_y$ denotes the $y$-th element of the $|\mathcal{Y}|$-dimensional probabilistic vector $D(E_t(x))$. In terms of parameterizing $r(x|s, y)$, a reconstructor $R : \mathbb{R}^K \times \mathcal{Y} \to \mathcal{X}$ is used, which takes concatenated $(s, y)$ as input and produce the corresponding reconstruction. Then, the *reconstruction* loss [7] is used to implement $-\mathbb{E}_{p_{\text{data}}(y|x)} \mathbb{E}_{q(s|x)} \log r(x|s, y)$,

$$\mathcal{L}_{\text{recon}}(R(E_s(x), y), x) = \|R(E_s(x), y) - x\|_2^2. \tag{80}$$

In terms of minimizing $I(S; T)$, we use a discriminator $d$ and an adversarial training manner,

$$\min_q \max_d \mathbb{E}_{q(s)q(t)} \log d(s, t) + \mathbb{E}_{q(s,t)} \log (1 - d(s, t)). \tag{81}$$

The discriminator $d$ is parameterized by $W : \mathbb{R}^{2K} \to \mathbb{R}$, which takes concatenated $(s, t)$ as input and produce the corresponding estimate of the probability that its input is a sample from $q(s, t)$ rather than from $q(s)q(t)$. As aforementioned in main text, we first sample from joint distribution $q(s, t)$ efficiently by first sampling $x$ from dataset uniformly at random and then sampling from $q(s, t|x)$ [8], followed by sampling from product of marginal distributions $q(s)q(t)$ by shuffling the samples from the joint distribution $q(s, t)$ along the batch axis [9].

During training process, an adversarial training algorithm is used, where the discriminator $W$ and the main architecture (*i.e.*, encoders $E_t$, $E_s$, decoder $D$ and reconstructor $R$) are trained jointly. In particular, the main architecture parameters are updated using the combination of the cross entropy loss, the reconstruction loss and the adversarial loss (see Eq. (82)). The discriminator is trained to correctly estimate of the probability that its input is a sample from the joint distribution $q(s, t)$ rather than from the product of marginals $q(s)q(t)$ (see Eq. (83)). Then whole training algorithm is summarized in Alg. 1.

**Algorithm 1** Disentangled information bottleneck.

**Require:** Training set $\mathcal{D} = \{x_i, y_i\}_{i=1}^{N}$, encoders $E_\text{t}, E_\text{s}$, decoder $D$ and reconstructor $R$, discriminator $W$, parameter $\theta$ of collection $\{E_\text{r}, E_\text{s}, D\}$, parameter $\phi$ of discriminator $W$, optimizers $g_\theta$ and $g_\phi$ for parameters $\theta$ and $\phi$ respectively

1: **while** not converge **do**
2:     Randomly select batch $\{x_i, y_i\}_{i \in \mathcal{B}}$
3:     Obtaining samples $\{t_i = E_\text{t}(x_i), s_i = E_\text{s}(x_i)\}_{i \in \mathcal{B}}$ from joint distribution $q(s, t)$
4:     Calculate loss $\mathcal{L}_\theta$ with respect to parameter $\theta$

$$\mathcal{L}_\theta = \frac{1}{|\mathcal{B}|} \sum_{i \in \mathcal{B}} \left( \mathcal{L}_\text{CE}(t_i, y_i) + \mathcal{L}_\text{recon}(R(s_i, y_i), x_i) - \log W(s_i, t_i) \right) \tag{82}$$

5:     Update parameter $\theta \leftarrow g_\theta \nabla_\theta \mathcal{L}_\theta$
6:     $\pi \leftarrow$ random shuffling on batch indices $\mathcal{B}$, obtaining samples $\{t_{\pi(i)}, s_{\pi(i)}\}_{i \in \mathcal{B}}$ from product of marginals $q(s) q(t)$
7:     Calculate loss $\mathcal{L}_\phi$ with respect to parameter $\phi$

$$\mathcal{L}_\phi = \frac{1}{|\mathcal{B}|} \sum_{i \in \mathcal{B}} -\log(1 - W(s_i, t_i)) - \log W\left(t_{\pi(i)}, s_{\pi(i)}\right) \tag{83}$$

8:     Update parameter $\phi \leftarrow g_\phi \nabla_\phi \mathcal{L}_\phi$
9: **end while**

## 3. Supplemental Experimental Results

Following [10], we use MLP models when compared to the IB Lagrangian baselines [10, 11, 12]. Following [10], for MNIST [13] and FashionMNIST [14] we directly use flattened images as input to the model while for CIFAR10 [15], Tiny-ImageNet [16] and SUN-RGBD [17] we use pretrained CNN models to extract the representations. Images in all datasets are normalized to $[0, 1]$ before training. All models are end-to-end trained from scratch with no additional dataset.

In the following sections, we compare different methods in terms of behavior on *IB Plane* (Section 3.1), generalization and adversarial robustness (Section 3.2) and out-of-distribution detection (Section 3.3) on FashionMNIST [14], CIFAR10 [15], Tiny-ImageNet [16] and SUN-RGBD [17] datasets. All the results are obtained in the same way as in the main text.

### 3.1. Behavior on *IB Plane*

In this section, we compare behaviors of different methods on *IB Plane*. The results are summarized in Figure 1-4. We have similar observations on FashionMNIST and CIFAR10 datasets as in the main text based on Figure 1 and Figure 2. In terms of Tiny-ImageNet and SUN-RGBD datasets, we observe that compression behaviors (*i.e.*, top-rows in Figure 3 and Figure 4) of IB Lagrangian variants are slightly different from that on FashionMNIST and CIFAR10 datasets. In particular, models trained on Tiny-ImageNet and SUN-RGBD datasets begin to compress at a relatively higher $\beta$. This is not only because both datasets contains natural images which are more challenging, but also because they have more categories (200 categories in Tiny-ImageNet and 19 categories in SUN-RGBD). Similarly, compression performances of our method on these two datasets are slightly inferior to that on FashionMNIST and CIFAR10 datasets. However, our method is still able to avoid prediction performance (*i.e.*, $I(T; Y)$) loss while efficiently compressing $I(X; T)$.

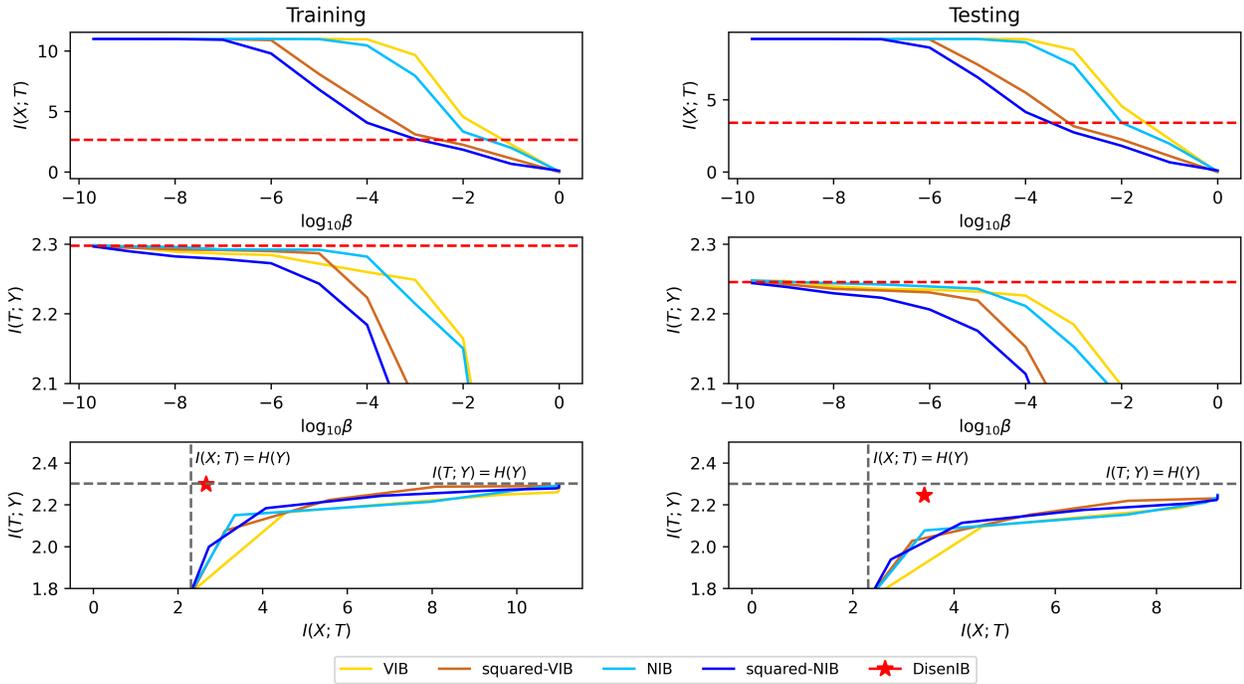

Figure 1. FashionMNIST training (**left**) and testing (**right**) results. **Top row:** $I(X;T)$ *v.s.* $\beta$ curve. **Middle row:** $I(T;Y)$ *v.s.* $\beta$ curve. **Bottom row:** IB-plane diagrams.

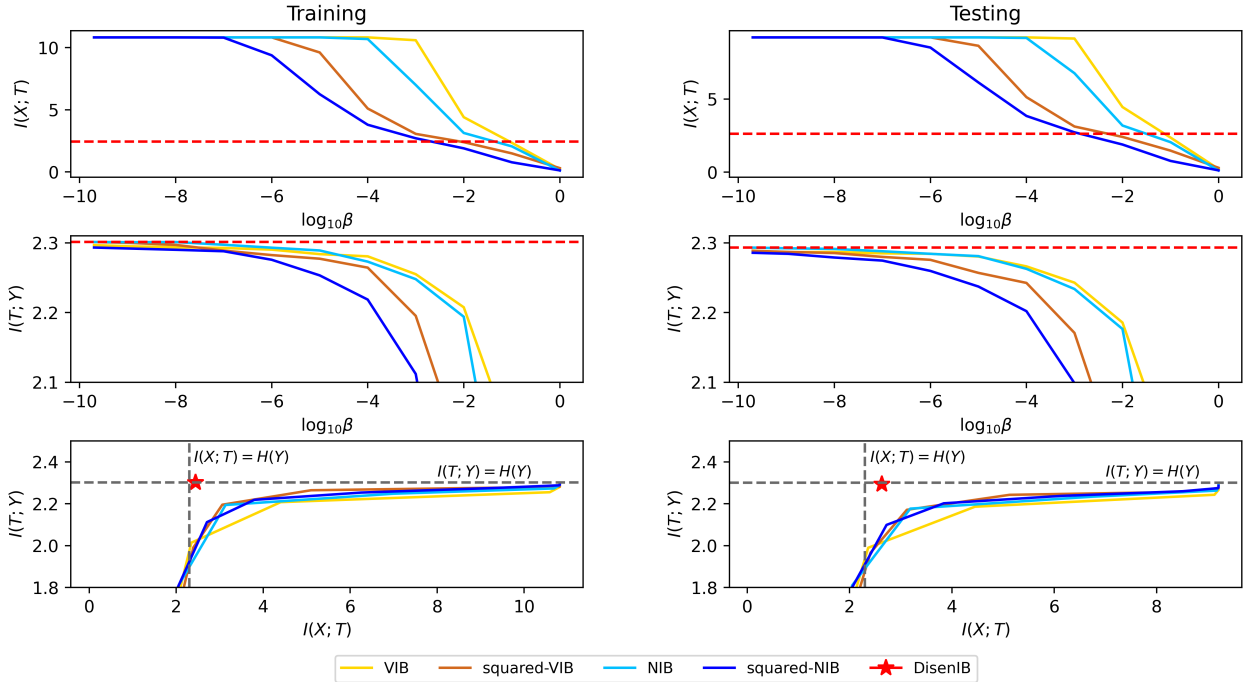

Figure 2. CIFAR10 training (**left**) and testing (**right**) results. **Top row:** $I(X;T)$ *v.s.* $\beta$ curve. **Middle row:** $I(T;Y)$ *v.s.* $\beta$ curve. **Bottom row:** IB-plane diagrams.

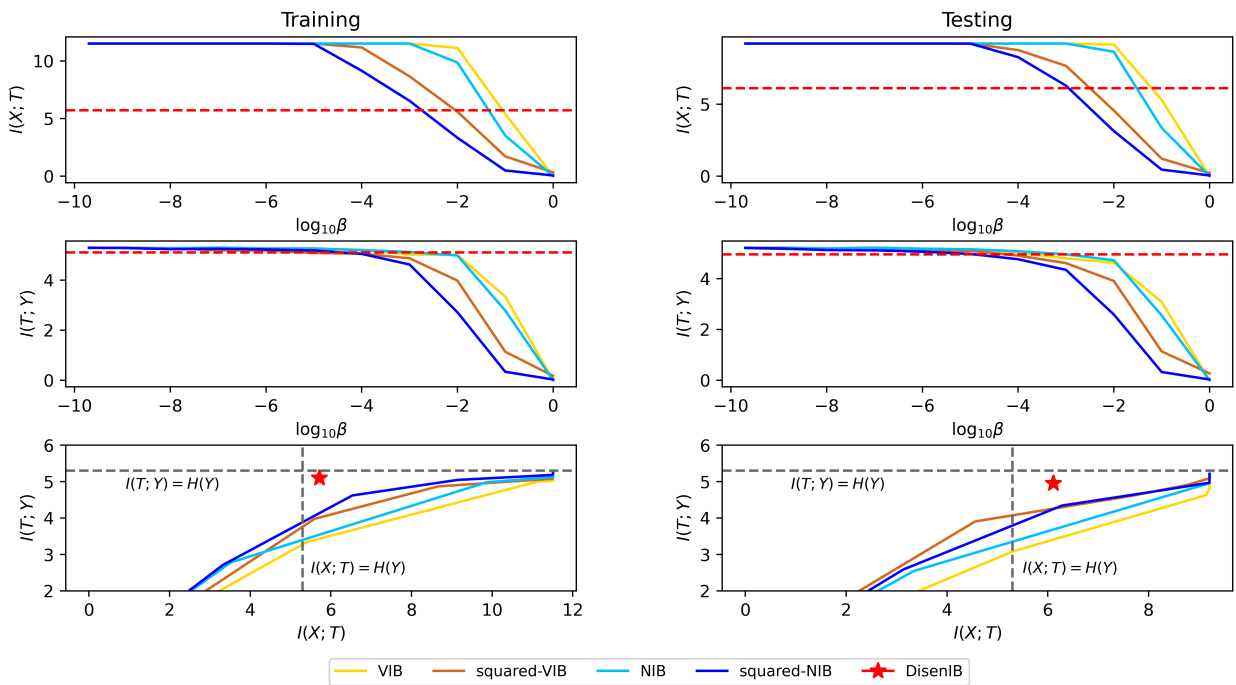

Figure 3. Tiny-ImageNet training (**left**) and testing (**right**) results. **Top row:** $I(X;T)$ *v.s.* $\beta$ curve. **Middle row:** $I(T;Y)$ *v.s.* $\beta$ curve. **Bottom row:** IB-plane diagrams.

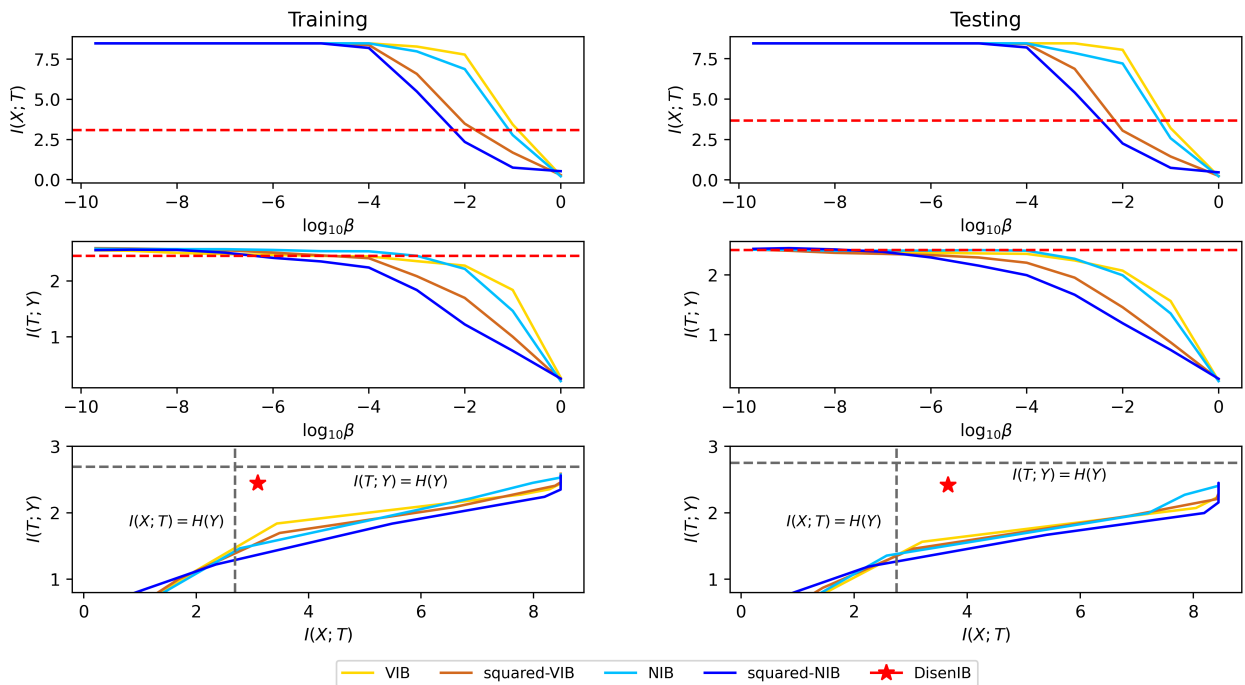

Figure 4. SUN-RGBD training (**left**) and testing (**right**) results. **Top row:** $I(X;T)$ *v.s.* $\beta$ curve. **Middle row:** $I(T;Y)$ *v.s.* $\beta$ curve. **Bottom row:** IB-plane diagrams.

| FashionMNIST | | Train | Test |
|---|---|---|---|
| | | VIB [10] / squared-VIB [12] / NIB [11] / squared-NIB [12] / DisenIB | |
| Generalization | | N/A | 90.1 / 89.5 / 89.7 / 89.6 / **90.1** |
| Adversary Robustness | $\epsilon = 0.1$ | 26.3 / 17.3 / 26.7 / 28.0 / **59.6** | 29.0 / 18.0 / 28.7 / 29.4 / **62.0** |
| | $\epsilon = 0.2$ | 3.7 / 4.0 / 2.6 / 0.7 / **45.5** | 4.4 / 4.1 / 0.3 / 0.8 / **48.4** |
| | $\epsilon = 0.2$ | 0.1 / 2.6 / 0.1 / 0.3 / **32.5** | 0.1 / 2.5 / 0.1 / 0.4 / **34.9** |
| CIFAR10 | | Train | Test |
| | | VIB [10] / squared-VIB [12] / NIB [11] / squared-NIB [12] / DisenIB | |
| Generalization | | N/A | 97.9 / 96.5 / 97.3 / 93.7 / **98.8** |
| Adversary Robustness | $\epsilon = 0.1$ | 64.3 / 41.1 / 69.2 / 62.3 / **70.4** | 62.4 / 42.3 / 65.2 / 61.0 / **70.3** |
| | $\epsilon = 0.2$ | 18.1 / 9.7 / 22.8 / 27.1 / **49.2** | 19.8 / 9.9 / 23.6 / 25.2 / **48.7** |
| | $\epsilon = 0.2$ | 2.0 / 6.9 / 5.2 / 10.1 / **29.3** | 3.2 / 7.1 / 4.3 / 10.7 / **30.0** |
| Tiny-ImageNet | | Train | Test |
| | | VIB [10] / squared-VIB [12] / NIB [11] / squared-NIB [12] / DisenIB | |
| Generalization | | N/A | 80.6 / 79.0 / 80.7 / 67.3 / **84.4** |
| Adversary Robustness | $\epsilon = 0.1$ | 42.3 / 31.5 / 47.2 / 41.3 / **57.3** | 42.1 / 30.7 / 48.2 / 40.0 / **51.2** |
| | $\epsilon = 0.2$ | 9.2 / 4.1 / 10.8 / 8.1 / **21.5** | 8.7 / 4.2 / 11.6 / 7.8 / **19.0** |
| | $\epsilon = 0.2$ | 1.0 / 0.7 / 2.2 / 1.3 / **7.1** | 0.6 / 1.3 / 3.1 / 0.9 / **5.8** |
| SUN-RGBD | | Train | Test |
| | | VIB [10] / squared-VIB [12] / NIB [11] / squared-NIB [12] / DisenIB | |
| Generalization | | N/A | 63.3 / 55.8 / 63.2 / 59.3 / **66.2** |
| Adversary Robustness | $\epsilon = 0.1$ | 37.1 / 25.1 / 41.2 / 35.3 / **41.3** | 36.4 / 23.7 / 42.2 / 32.0 / **40.2** |
| | $\epsilon = 0.2$ | 5.7 / 3.7 / 7.8 / 4.1 / **12.5** | 4.8 / 2.2 / 8.6 / 3.5 / **12.0** |
| | $\epsilon = 0.2$ | 0.5 / 0.1 / 1.2 / 0.3 / **3.4** | 1.2 / 0.1 / 1.4 / 0.9 / **2.8** |

Table 1. Classification mean accuracy and adversarial robustness (%) on FashionMNIST, CIFAR10, Tiny-ImageNet and SUN-RGBD datasets.

## 3.2. Generalization and Adversarial Robustness

In this section, we compare generalization performance and adversarial robustness of different methods on FashionMNIST [14], CIFAR10 [15], Tiny-ImageNet [16] and SUN-RGBD [17] datasets. The results are summarized in Table 1, which are obtained in the same way as in the main text. Based on Table 1, we have similar observations as in the main text. Specifically, since a reduced information (*i.e.*, $I(T; Y)$) will degenerate a model's generalization performance and adversarial robustness, our method outperforms existing ones due to the avoidance of information reduction, especially in terms of adversarial robustness on natural images datasets Tiny-ImageNet and SUN-RGBD. As aforementioned, since these two natural images datasets contains challenging real-world images within much more categories, it is difficult for the model to learn representations that generalizes well, hence a compressed model without prediction performance loss especially performs well in terms of generalization and adversarial robustness [18, 19].

| FashionMNIST | SUN-RGBD [17] | Gaussian Noise |
| --- | --- | --- |
| | VIB [10] / squared-VIB [12] / NIB [11] / squared-NIB [12] / DisenIB | |
| FPR (95% TPR) ↓ | 27.8 / 27.7 / 18.5 / 20.5 / **0.1** | 18.8 / 15.5 / 8.4 / 14.9 / **0.2** |
| AUROC ↑ | 84.5 / 77.9 / 89.9 / 90.4 / **98.6** | 89.7 / 79.7 / 92.2 / 91.4 / **99.1** |
| AUPR In ↑ | 85.3 / 77.1 / 90.5 / 91.6 / **98.8** | 90.3 / 79.1 / 92.5 / 92.3 / **99.1** |
| AUPR Out ↑ | 81.0 / 75.0 / 88.4 / 88.4 / **98.1** | 88.3 / 76.7 / 91.4 / 90.0 / **98.8** |
| Detection Error ↓ | 22.7 / 28.0 / 17.2 / 17.2 / **0.5** | 17.7 / 26.7 / 15.1 / 16.0 / **0.3** |
| CIFAR10 | SUN-RGBD [17] | Gaussian Noise |
| | VIB [10] / squared-VIB [12] / NIB [11] / squared-NIB [12] / DisenIB | |
| FPR (95% TPR) ↓ | 29.1 / 20.5 / 19.0 / 22.1 / **1.3** | 25.6 / 25.4 / 14.4 / 27.0 / **2.7** |
| AUROC ↑ | 68.7 / 59.9 / 67.5 / 63.5 / **87.4** | 72.7 / 78.8 / 72.9 / 80.0 / **89.7** |
| AUPR In ↑ | 63.4 / 52.0 / 60.8 / 55.5 / **85.6** | 73.3 / 71.5 / 76.8 / 74.9 / **85.8** |
| AUPR Out ↑ | 69.5 / 64.7 / 68.7 / 65.8 / **87.9** | 76.1 / 79.2 / 73.6 / 79.6 / **89.5** |
| Detection Error ↓ | 35.4 / 38.4 / 35.5 / 37.5 / **7.7** | 29.9 / 26.2 / 28.9 / 25.7 / **5.9** |
| Tiny-ImageNet | CIFAR10 [15] | Gaussian Noise |
| | VIB [10] / squared-VIB [12] / NIB [11] / squared-NIB [12] / DisenIB | |
| FPR (95% TPR) ↓ | 23.6 / 23.7 / 26.5 / 26.8 / **20.3** | 26.4 / 19.1 / 28.1 / 25.9 / **15.1** |
| AUROC ↑ | 87.9 / 83.1 / 85.1 / 82.9 / **89.3** | 84.5 / 86.4 / 85.6 / 89.9 / **93.9** |
| AUPR In ↑ | 90.5 / 85.9 / 86.3 / 86.9 / **92.2** | 81.2 / 82.9 / 86.8 / 82.3 / **95.8** |
| AUPR Out ↑ | 84.3 / 79.6 / 83.1 / 82.2 / **87.8** | 86.2 / 80.0 / 84.7 / 86.6 / **89.9** |
| Detection Error ↓ | 19.1 / 24.2 / 20.2 / 21.2 / **8.6** | 16.1 / 19.4 / 19.6 / 16.3 / **9.3** |
| SUN-RGBD | CIFAR10 [15] | Gaussian Noise |
| | VIB [10] / squared-VIB [12] / NIB [11] / squared-NIB [12] / DisenIB | |
| FPR (95% TPR) ↓ | 20.4 / 19.9 / 25.4 / 28.5 / **17.1** | 18.5 / 19.7 / 16.4 / 21.7 / **10.1** |
| AUROC ↑ | 82.6 / 81.6 / 81.2 / 76.6 / **89.5** | 86.8 / 85.5 / 87.4 / 82.8 / **91.5** |
| AUPR In ↑ | 83.8 / 79.5 / 82.2 / 74.3 / **88.6** | 82.8 / 81.6 / 88.8 / 79.4 / **89.6** |
| AUPR Out ↑ | 81.7 / 80.2 / 89.8 / 78.1 / **92.9** | 84.5 / 86.5 / 87.2 / 78.8 / **90.4** |
| Detection Error ↓ | 15.5 / 14.0 / 12.9 / 16.0 / **12.7** | 14.7 / 15.7 / 13.6 / 16.7 / **11.2** |

Table 2. Distinguishing in- and out-of-distribution testing data for FashionMNIST, CIFAR10, Tiny-ImageNet and SUN-RGBD image classification (%). ↑ indicates larger value is better, and ↓ indicates lower value is better.

## 3.3. Out-of-distribution Detection

In this section, we compare performance on out-of-distribution detection of different methods on FashionMNIST [14], CIFAR10 [15], Tiny-ImageNet [16] and SUN-RGBD [17] datasets. The results are summarized in Table 2, which are obtained in the same way as in the main text. Following main text, at detection time, examples from corresponding test set can be viewed as in-distribution data and we use examples from scene-centric natural image dataset SUN-RGBD [17] out-distribution data for FashionMNIST and CIFAR10 benchmark datasets. In terms of Tiny-ImageNet and SUN-RGBD datasets, we use examples from CIFAR10 as out-distribution data, considering the large distribution gaps between them. Based on Table 2, we have similar observations as in the main text. While we observe that performance of our method on real-world Tiny-ImageNet and SUN-RGBD datasets are slightly inferior to that on benchmark datasets, our method is still able to outperform existing methods. This is because results of existing methods are obtained at the same compression level as ours, the bottleneck information is reduced due to the trade-off involved in the IB Lagrangian, making it difficult to learn a generalized model with good capacity on detecting out-of-distribution data [20, 18]. Since our method avoids information reduction while performing efficiently compression, performances of our method are consistently superior than existing ones.



\end{document}